\documentclass{article}
\usepackage{officetown_arxiv,times}
\usepackage[T1]{fontenc}
\usepackage{graphicx}
\usepackage{float}
\usepackage{wrapfig}
\usepackage{hyperref}
\usepackage{url}
\usepackage{booktabs,array,tabularx,longtable,multirow}
\usepackage{xurl}
\usepackage{placeins}
\newcommand{\NA}{\textemdash}

\newcolumntype{Y}{>{\raggedright\arraybackslash}X}
\newcolumntype{R}[1]{>{\raggedleft\arraybackslash}p{#1}}
\usepackage{colortbl}
\definecolor{OTTableInk}{HTML}{38536B}
\definecolor{OTTableMuted}{HTML}{788591}
\definecolor{OTTableHeader}{HTML}{EDF1F5}
\definecolor{OTTableGroup}{HTML}{F3F5F7}
\definecolor{OTTableFocus}{HTML}{EDF4F1}
\definecolor{OTRankOne}{HTML}{AACCB9}
\definecolor{OTRankTwo}{HTML}{CFDFD6}
\definecolor{OTRankThree}{HTML}{EBF2EE}
\definecolor{OTPassOne}{HTML}{ADC6DE}
\definecolor{OTPassTwo}{HTML}{D0DEEC}
\definecolor{OTPassThree}{HTML}{EAF0F7}
\newcommand{\OTRank}[2]{\cellcolor{#1}#2}
\definecolor{OTSecondary}{HTML}{657483}
\definecolor{OTBase}{HTML}{7B8794}
\definecolor{OTSFT}{HTML}{507DA3}
\definecolor{OTOPD}{HTML}{307C65}
\definecolor{OTSFTGain}{HTML}{EEF3F8}
\definecolor{OTOPDGain}{HTML}{EAF3EF}

\newcommand{\OTStage}[2]{\textcolor{#1}{\rule{0.45em}{0.45em}} #2}

\newcommand{\OTResultSetup}{\renewcommand{\NA}{\textcolor{OTTableMuted}{\textemdash}}}

\newcommand{\OTBenchmarkTasks}{406}
\newcommand{\OTSFTTrajectories}{2,202}

\newcommand{\OTRubricCriteria}{4,836}
\newcommand{\OTOPDTasks}{828}
\newcommand{\JuniperInitialFiles}{27}
\newcommand{\JuniperOutputFiles}{470}
\newcommand{\JuniperTotalFiles}{497}
\newcommand{\JuniperDaySixtyFiles}{209}
\newcommand{\JuniperDayOneTwentyFiles}{376}

\newcommand{\papertitle}{OfficeTown: Generative Organizations for Evolving Agent Intelligence}
\newcommand{\paperkeywords}{Data Synthesis, Agent Benchmarking, Organizational Simulation}
\title{\papertitle}
\author{%
\normalfont
Hongchen Wei$^{1,\dagger,\ddagger}$, Daohong Ying$^{2,\dagger,\ddagger}$, Zhi Hong$^{3,\dagger,\ddagger}$, Aidan Wu$^{4,\dagger,\ddagger}$, 
Bei Liu$^{5,\ast}$, Ruichun Ma$^5$ \\[0.15em] Yifan Yang$^5$, Qi Dai$^5$, Kai Qiu$^5$, 
Rui Xu$^6$, Dongdong Chen$^6$, Zhenzhong Chen$^1$, Chong Luo$^5$\\[0.4em]
$^1$ Wuhan University \quad $^2$ University of California, San Diego\\
$^3$ South China University of Technology \quad $^4$ Cornell University\\
$^5$ Microsoft Research Asia \quad $^6$ Microsoft\\[0.25em]
\small
$^\dagger$ Equal contribution. \quad $^\ddagger$ Work done during an internship at MSRA. \quad $^\ast$ Project leader.
}
\hypersetup{
  pdftitle={\papertitle},
  pdfauthor={Hongchen Wei, Daohong Ying, Zhi Hong, Aidan Wu, Bei Liu, Ruichun Ma, Yifan Yang, Qi Dai, Kai Qiu, Rui Xu, Dongdong Chen, Zhenzhong Chen, Chong Luo},
  pdfkeywords={\paperkeywords}
}

\begin{document}
\addtocontents{toc}{\protect\setcounter{tocdepth}{-1}}
\maketitle
\fancyhead{}

\begin{abstract}
AI agents increasingly use files and software tools to produce professional
deliverables. Such work draws on materials, decisions, and feedback accumulated across
assignments. We introduce OfficeTown, a generative simulation framework for professional
organizations in which tasks and their context arise from continuing organizational
work. From a natural-language description of an organization's purpose and activities,
OfficeTown initializes an organization whose members coordinate work, delegate bounded assignments
to assistant agents, and review their deliverables. Artifacts, decisions, and
unresolved matters persist and shape subsequent work, while responsibilities and
permissions determine what each member can see and do. We curate work requests, their
context, and review feedback into executable resources for evaluation and learning. We
construct OfficeTown Bench with 406 tasks and 4,836 evaluation criteria from 23
simulated organizations across eight professional domains, incorporating domain-expert
feedback. Frontier-model evaluations reveal substantial remaining headroom. Supervised
fine-tuning and on-policy distillation on OfficeTown data substantially improve
Qwen3.5-9B on external professional-work benchmarks, including GDPval. Supplementary
evaluations show that the principal--assistant harness achieves higher scores than the
single- and multi-agent baselines tested. These results highlight the value of
organizational simulation in generating work-grounded data and environments for the
continued development of generalist agents.
\par\noindent\textbf{Project homepage:}\enspace \url{https://officeintelligence.github.io/}
\end{abstract}

\begin{figure}[!t]
  \centering
  \includegraphics[width=\linewidth]{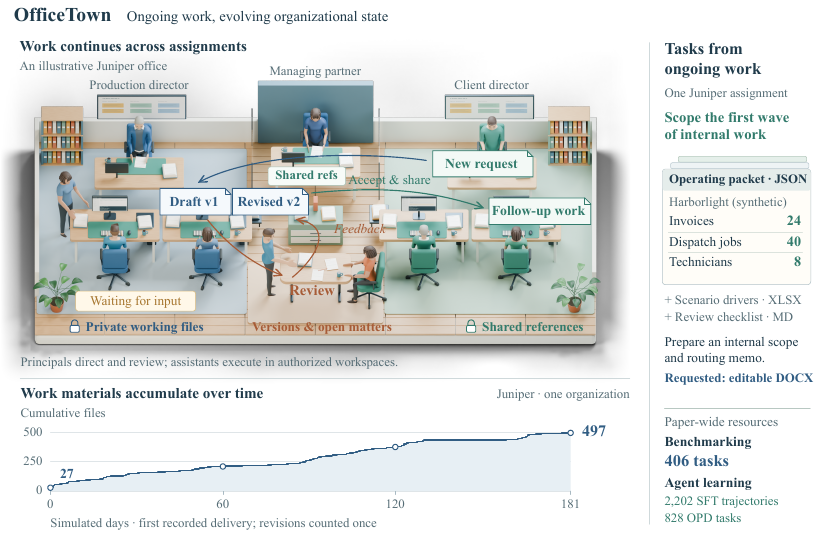}
  \caption{OfficeTown models ongoing work in a persistent, permissioned organization.
  Professionals coordinate requests and review deliverables; assistants execute
  within authorized workspaces. Private drafts, revisions and unresolved matters
  carry context into subsequent work. The ten-role Juniper scene and its work
  links are illustrative. The lower plot shows cumulative
  materials from one Juniper organization over 181 simulated days (not wall-clock
  time): 27 initial files and 470 subsequent logical outputs, with revisions and
  exact copies deduplicated (Appendix~\ref{sec:longitudinal-analysis}). These
  counts track material accumulation, not task quality or success. The right-hand
  panel shows synthetic inputs and the request for a retained scope-and-routing
  memo, not a model response.}
  \label{fig:officetown-generation}
\end{figure}

\section{Introduction}
\label{sec:introduction}

Large language models are increasingly expected to produce professional deliverables using files, software tools, and organizational knowledge. Recent benchmarks~\citep{patwardhan2025gdpval} and
environment-synthesis methods~\citep{dong2026agentworld} provide valuable resources for evaluating these capabilities, but typically treat tasks as isolated instances.
Real professional work is rarely isolated. In organizations, assignments arise from ongoing activity rather than appearing independently. Deliverables are reviewed and revised, decisions create follow-up requests, and materials accumulated in one project become inputs to later work. Tasks therefore emerge from a continuing process in which work products and work context co-evolve.

This observation suggests a different perspective. Instead of generating standalone tasks, we may seek to model the system that continuously produces professional work. Such a system would naturally generate new assignments, contextual materials, review feedback, and long-range dependencies, offering a scalable source of both evaluation and training data.

Professional organizations are precisely such systems. We therefore introduce OfficeTown, a generative simulation framework for professional organizations. From a natural-language description of an organization's purpose and activities, OfficeTown constructs persistent organizational structures, roles, resources, and authority boundaries, then simulates ongoing work as professionals coordinate, delegate, review, and follow up on assignments. Organizational state is persistent but not globally observable:
each member has a permission-filtered view determined by role, ownership, and sharing
rules, and acts within their role-specific authority. Figure~\ref{fig:officetown-generation} illustrates a Juniper instance.

These professionals act as principals who coordinate work, delegate bounded assignments
to assistant agents, and review their deliverables. Assistants execute within authorized workspaces. Artifacts and decisions update organizational state, while
unresolved matters prompt follow-up work. We package selected assignments with their
task-time context into executable resources for evaluation and learning.
Figure~\ref{fig:officetown-generation} (right) shows a concrete assignment with
supporting materials and a requested internal scope and routing memo.

We construct OfficeTown Bench with 406 tasks and 4,836 evaluation criteria from 23
simulated organizations across eight professional domains, incorporating domain-expert
feedback into task and rubric refinement. Delegating professionals span 139 distinct
job titles. Training resources comprise 2,202 demonstrations for supervised fine-tuning
(SFT) and 828 task--environment pairs for on-policy distillation (OPD).

Frontier-model evaluations reveal substantial remaining headroom on OfficeTown Bench.
Training on OfficeTown data improves Qwen3.5-9B on external professional-work
benchmarks. On 79 shared GDPval tasks, its rubric score rises from 65.75 to 73.65 after
SFT and OPD. Supplementary evaluations favor the principal--assistant harness
over tested single- and multi-agent baselines.

Our contributions are:
\begin{itemize}
  \item We introduce a generative simulation framework for professional organizations in which
persistent, permissioned state shapes subsequent work.
  \item We construct an expert-informed professional-work benchmark and training resources from
ongoing organizational work.
  \item We demonstrate that the resulting tasks challenge frontier models and that OfficeTown
data improves performance on external benchmarks.
\end{itemize}
\suppressfloats[t]
\section{OfficeTown: A Generative Organizational Simulation Framework}
\label{sec:framework}

\begin{figure}[t]
  \centering
  \includegraphics[width=\linewidth]{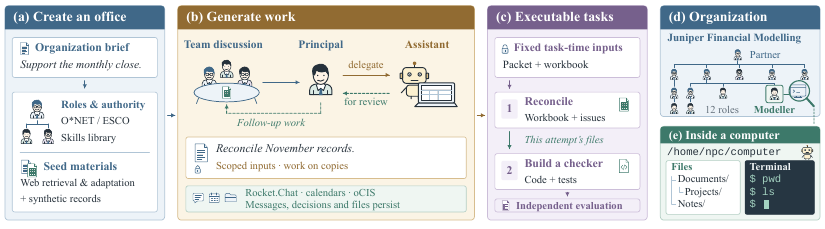}
  \caption{OfficeTown's organizational simulation.
  \textbf{(a)} Organizational objectives, roles, and seed materials establish an organization.
  \textbf{(b)} Coordination, delegation, execution, and review connect ongoing work
  through retained materials and follow-up assignments.
  \textbf{(c)} Tasks drawn from this activity provide fixed inputs and executable
  environments for training and evaluation.
  \textbf{(d)} The recorded twelve-role hierarchy of Juniper Financial Modelling;
  the senior modeller's computer is highlighted.
  \textbf{(e)} Filesystem and terminal views used by an assistant.
  The reconciliation and Juniper examples are separate.
  Directory names are verified; interfaces and terminal activity are schematic.}
  \label{fig:officetown-framework}
\end{figure}

OfficeTown initializes an organization with roles, goals, resources, and authority, then
simulates ongoing professional work progressing asynchronously
(Figure~\ref{fig:officetown-framework}). Requests, messages, commitments, and unresolved
matters activate professionals, who coordinate work and delegate bounded assignments to
assistants. They review delivered artifacts, request revisions, and arrange follow-up
work. Files, communications, and decisions update persistent, permissioned
organizational state throughout this process, shaping subsequent assignments.

\subsection{Organizations, Roles, and Authority}
\label{sec:simulated-organizations}

An OfficeTown instance represents a professional organization or team initialized from
a natural-language description of its purpose and activities
(Figure~\ref{fig:officetown-framework}a). The description guides organizational objectives,
professional roles, and reporting relationships. Each role defines continuing
responsibilities and role-specific authority: which matters the professional handles and what they may execute,
delegate, approve, or share. Figure~\ref{fig:officetown-framework}d shows Juniper
Financial Modelling, a twelve-role instance.

Each role is represented by a model-driven professional whose profile draws on O*NET and
ESCO occupational descriptions. Professionals act as principals responsible for ongoing
work across assignments; assistants perform delegated executions.

The organization starts with documents and data, termed the seed. These combine
retrieved or adapted public sources with organization-specific synthetic records,
preserving source references and the distinction between these origins. The seed
provides initial working context that subsequent work can extend or revise.
Appendices~\ref{app:organization-construction}--\ref{app:example-configuration}
detail construction, role skills, and Juniper's setup.

\subsection{Persistent and Permissioned Organizational State}
\label{sec:persistent-state}

Organizational state includes personal working files, shared materials, communications,
commitments, and pending decisions. Each professional maintains a persistent workspace
(Figure~\ref{fig:officetown-framework}e) and structured work records separate from
active conversation context, so refreshing a model session does not erase unresolved
work. Each activation presents selected work records alongside current requests,
prioritizing outstanding commitments and discussions awaiting a response.

This state is persistent but not globally observable. Role permissions, ownership,
sharing rules, and confidentiality determine each member's information view;
role-specific authority determines what they may do with it. Reporting relationships do not
automatically grant file access. Assistants receive writable working files and
permission-filtered, read-only shared references within the principal's authorization,
not access to other professionals' private workspaces.

Saving, accepting, and sharing a deliverable are distinct actions. Saving retains files,
including drafts and pending revisions; professional review determines acceptance;
explicit sharing makes an accepted deliverable available to authorized recipients. Thus,
saved work need not be accepted or visible to others. Acceptance, version,
classification, and recipient permissions are checked before broader publication.
Appendices~\ref{app:sessions-memory}--\ref{app:execution-recovery} cover workspace
checks, sessions, and recovery.

\subsection{Coordination, Delegation, and Review}
\label{sec:task-generation}

Work follows a shared simulated calendar, with professionals activated asynchronously by
incoming requests, messages, meetings, follow-ups, or assistant deliveries
(Figure~\ref{fig:officetown-framework}b). The simulation controls activation and
permitted actions; the professional's model decides how to respond using current
requests, commitments, and accessible materials. Several matters can therefore progress
concurrently. Appendix~\ref{app:execution-recovery} details scheduling and clock control.

When responsibilities intersect, members exchange permitted materials, question
assumptions, and seek colleagues' input. Team leads can meet to clarify objectives,
identify missing inputs, and allocate responsibilities; meetings are not required for
every assignment. Agreed work is recorded as requests specifying the intended outcome,
responsible person, supporting materials, and deadline. Appendix~\ref{app:communication}
provides a recorded example.

For delegated execution, the principal specifies the objective, available inputs,
constraints, and expected deliverable. The assistant works within the authorized
workspace and can ask the principal to clarify requirements or consult colleagues. On
delivery, the principal inspects the actual artifact and decides whether to accept it,
request revisions, or seek further information. Review can change both deliverable
requirements and the work that other members need to undertake. These reviews are
organizational decisions within the simulated organization, not independent benchmark
judgments.

Revisions build on submitted artifacts and review comments. Accepted results can become
inputs to authorized colleagues' work, while unresolved questions remain available for
later investigation. These updates link assignments through accumulated context,
extending work beyond the initial seed.
Appendix~\ref{sec:longitudinal-analysis} illustrates material accumulation over six
simulated months.

\subsection{From Ongoing Work to Executable Tasks}
\label{sec:task-environments}

An executable task pairs a work request with materials available to its responsible
professional when the work was assigned (Figure~\ref{fig:officetown-framework}c). This
task-time snapshot preserves both temporal and permission boundaries: personal files,
authorized shared resources, and existing artifacts needed for revision may be included,
but later solutions and evaluator-only information are excluded.

Each attempt starts in a fresh workspace initialized from these fixed inputs, where the
agent can inspect and edit files, execute commands, and use the provided interfaces. The
continuing organization is the source of the assignment; individual evaluation or
learning attempts do not require replaying its entire history. Section~\ref{sec:data}
explains selection, expert-informed refinement, and packaging into benchmark and
training resources.
\section{OfficeTown Data and Benchmark}
\label{sec:data}

We curate assignments and their task-time environments from ongoing organizational work
into OfficeTown Bench and training resources. The benchmark measures professional task
completion, while execution demonstrations and interactive task environments support
learning.

\subsection{Benchmark Design}
\label{sec:dataset-overview}

\begin{figure}[!t]
  \centering
  \includegraphics[width=\linewidth]{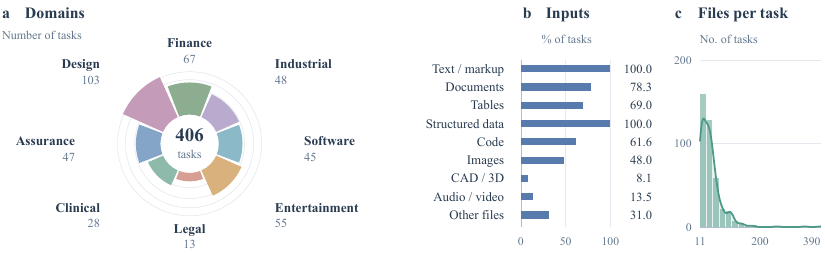}
  \caption{OfficeTown Bench composition. (a) Domains, with petal areas and labels
  indicating task counts.
  (b) Input-type coverage, including archive contents; types overlap.
  (c) Input-file counts: 20-file bins starting at 11, with a smoothed frequency curve.
  All panels use all \OTBenchmarkTasks{} tasks; an archive counts as one file in (c).}
  \label{fig:officetown-data}
  \par
\noindent
\begin{minipage}[t]{0.48\linewidth}

\begingroup
\makeatletter\def\@captype{table}\makeatother
\caption{OfficeTown Bench statistics.
Per-task rows: mean / median. Total counts file instances across
tasks; unique counts deduplicate by content. Archives count as one file.}
\label{tab:officetown-resources}

\centering

\begin{center}
\begin{tabularx}{\linewidth}{Yr}
\toprule
Statistic & Value \\
\midrule
Tasks & 406 \\
Professional domains & 8 \\
Source organizations & 23 \\
Distinct role titles & 139 \\
\midrule
Input files (task-wise total) & 18,082 \\
Unique input files & 6,951 \\
Input files / task & 44.54 / 37 \\
Input types / task & 5.10 / 5 \\
\midrule
Rubric criteria (total) & 4,836 \\
Rubric criteria / task & 11.91 / 11 \\
Contributing domain experts & 71 \\
\bottomrule
\end{tabularx}
\end{center}
\par
\endgroup
\par

\end{minipage}\hfill
\begin{minipage}[t]{0.48\linewidth}

\centering
\caption{An OfficeTown Bench task. Dashed files are requested outputs;
the rubric is evaluator-only.}
\label{fig:officetown-task-example}
\includegraphics[width=\linewidth]{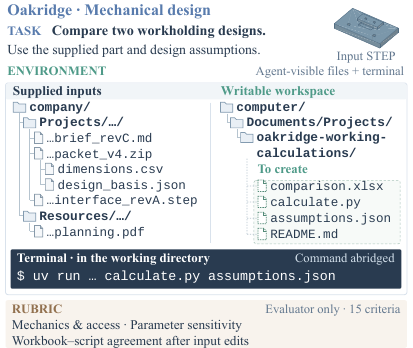}
\end{minipage}
\par
\end{figure}

OfficeTown Bench covers eight professional domains selected for economic relevance and
digital executability, inspired by GDPval~\citep{patwardhan2025gdpval}. It contains
\OTBenchmarkTasks{} tasks and \OTRubricCriteria{} evaluation criteria from 23 simulated
organizations, involving delegating professionals with 139 distinct job titles. The
benchmark includes 247 file-based tasks and 159 tasks involving service state.
Figure~\ref{fig:officetown-data} and Table~\ref{tab:officetown-resources} report its
domain coverage and input-material composition.

Each sample combines a work request, an executable environment, and a task-specific
rubric. Environments supply the files, tools, and authorized organizational state for the
assignment; rubrics assess professional requirements but remain outside the agent's
starting context.

Figure~\ref{fig:officetown-task-example} illustrates a mechanical-design assignment.
Given workpiece geometry, feature tables, design assumptions, and engineering
references, the agent must compare two locating-and-clamping approaches in an editable
workbook with reproducible calculation code. The rubric assesses mechanical
plausibility, sensitivity to uncertain inputs, and agreement between the workbook and
code.

\subsection{Construction and Expert Review}
\label{sec:officetown-bench}

We inspect delegated requests and their working materials, then edit them into
self-contained evaluation tasks while retaining work objectives, constraints, and
intended deliverables. Task environments preserve the relevant starting materials,
including artifacts requiring revision, but exclude historical solutions to the
evaluated task and evaluator-only information.

Rubrics are drafted by a language model from task requirements and inspected source
materials. Human-authored examples guide criterion formulation without supplying
task-specific facts, thresholds, or answers. A separate model-based review checks
professional correctness, missing coverage, overlapping criteria, and unsupported
restrictions. Revisions preserve professional requirements.

During benchmark construction, 71 domain experts contributed feedback on task realism,
clarity, input sufficiency, and feasibility, as well as rubric correctness, coverage,
and alignment. Expert assignment follows relevant specialization, occupational
experience, and familiarity with working materials and applicable standards. Experts
propose edits, removals, merges, or splits and rate importance as Critical (P0), Major
(P1), Minor (P2), or Optional (P3). These annotations concern tasks and rubrics, not
individual agent answers. The released task and criterion sets are fixed for model
comparisons. Appendix~\ref{app:data} details construction, annotation instructions, and
review coverage.

\subsection{Training Data}
\label{sec:training-data}

The SFT resource comprises \OTSFTTrajectories{} complete execution trajectories
distilled from DeepSeek-V4-Flash, pairing task inputs with teacher reasoning, tool
calls, observations, and final responses. We retain complete histories without
truncation or filtering by task-success score.

The OPD resource provides \OTOPDTasks{} task--environment pairs for the subsequent
distillation stage. Each supplies a work request and its initial environment without a
teacher response history, using a system message and tool interface aligned with SFT.
Appendix~\ref{app:training} describes resource construction;
Section~\ref{sec:experiments} reports training and evaluation results.
\begin{table}[!t]
\caption{OfficeTown Bench Pass@1 (\%) on \OTBenchmarkTasks{} tasks, with high reasoning effort.
Domain abbreviations follow Figure~\ref{fig:officetown-data}.
\colorbox{OTPassOne}{1st} \colorbox{OTPassTwo}{2nd}
\colorbox{OTPassThree}{3rd} mark within-column ranks at displayed precision;
ties share rank and skip positions.
Weighted Scores and scoring definitions appear in
Appendices~\ref{app:weighted-scores} and~\ref{app:data-quality}.}
\label{tab:otbench-fill}
\begingroup
\begin{center}
\begin{tabularx}{\linewidth}{Yrrrrrrrrr}
\toprule
\textbf{Model} & Overall & Assur. & Clin. & Des. & Ent. & Fin. & Ind. & Legal & Soft. \\
\midrule
GPT-6 Astra & \OTRank{OTPassOne}{74.4} & \OTRank{OTPassTwo}{63.8} & \OTRank{OTPassOne}{92.9} & \OTRank{OTPassOne}{56.3} & \OTRank{OTPassOne}{72.7} & \OTRank{OTPassOne}{88.1} & \OTRank{OTPassOne}{81.3} & \OTRank{OTPassOne}{92.3} & \OTRank{OTPassOne}{84.4} \\
GPT-5.6 Sol & 33.5 & 38.3 & 35.7 & 26.2 & 45.5 & 34.3 & 29.2 & \OTRank{OTPassThree}{46.2} & 28.9 \\
GPT-5.6 Terra & 39.4 & 48.9 & 32.1 & 35.0 & 43.6 & 40.3 & 43.8 & 38.5 & 33.3 \\
GPT-5.6 Luna & 32.5 & 48.9 & 53.6 & 21.4 & 41.8 & 31.3 & 22.9 & 30.8 & 28.9 \\
GPT-5.4 & 30.5 & 31.9 & 39.3 & 25.2 & 41.8 & 32.8 & 29.2 & 30.8 & 20.0 \\
GPT-5.2 & 21.2 & 27.7 & 28.6 & 13.6 & 27.3 & 31.3 & 14.6 & 15.4 & 13.3 \\
Claude Opus 5.5 & \OTRank{OTPassThree}{50.7} & \OTRank{OTPassThree}{55.3} & \OTRank{OTPassTwo}{67.9} & \OTRank{OTPassThree}{38.8} & \OTRank{OTPassTwo}{60.0} & \OTRank{OTPassThree}{49.3} & \OTRank{OTPassThree}{50.0} & \OTRank{OTPassTwo}{61.5} & \OTRank{OTPassThree}{51.1} \\
DeepSeek V4-Pro & 18.5 & 23.4 & 14.3 & 14.6 & 30.9 & 19.4 & 12.5 & 0.0 & 20.0 \\
DeepSeek V4-Flash & 13.5 & 17.0 & 10.7 & 11.7 & 23.6 & 10.4 & 10.4 & 15.4 & 11.1 \\
Kimi-K2.5 & 4.9 & 10.6 & 7.1 & 1.0 & 14.5 & 1.5 & 2.1 & 7.7 & 2.2 \\
Grok-4.7 & \OTRank{OTPassTwo}{58.4} & \OTRank{OTPassOne}{70.2} & \OTRank{OTPassThree}{57.1} & \OTRank{OTPassTwo}{54.4} & \OTRank{OTPassThree}{47.3} & \OTRank{OTPassTwo}{65.7} & \OTRank{OTPassTwo}{58.3} & \OTRank{OTPassThree}{46.2} & \OTRank{OTPassTwo}{62.2} \\
Grok-4.3 & 8.1 & 19.1 & 14.3 & 6.8 & 12.7 & 4.5 & 0.0 & 7.7 & 4.4 \\
Mistral-Large-3 & 1.2 & 2.1 & 0.0 & 0.0 & 7.3 & 0.0 & 0.0 & 0.0 & 0.0 \\
\bottomrule
\end{tabularx}
\end{center}
\endgroup

\end{table}
\begin{figure}[!ht]
\centering
\includegraphics[width=\linewidth]{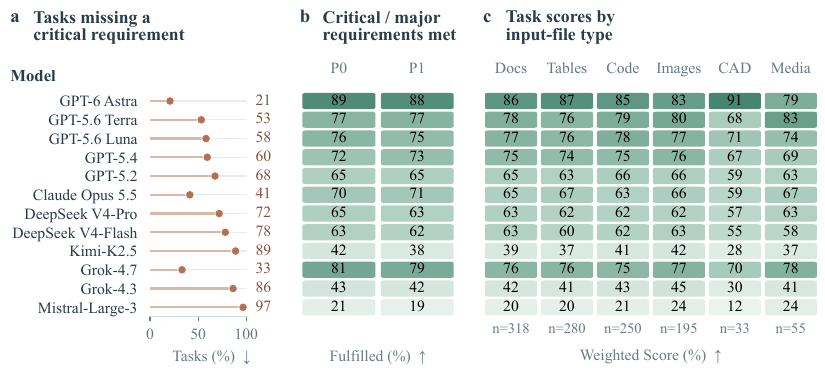}
\caption{Professional requirements beyond aggregate scores.
(a) Judged tasks missing at least one applicable critical (P0) requirement.
(b) Task-averaged fulfillment of critical and major requirements (P0/P1).
(c) Mean task scores grouped by supplied input-file types; groups overlap.
Twelve models with bound item-level judgments are shown. Item-level panels exclude
unjudged outcomes; file-group scores retain the main table's declared zeros.
Definitions and full values are in Appendix~\ref{app:diagnostics}.}
\label{fig:bench-diagnostics}
\end{figure}

\begin{figure}[!t]
\centering
\includegraphics[width=\linewidth]{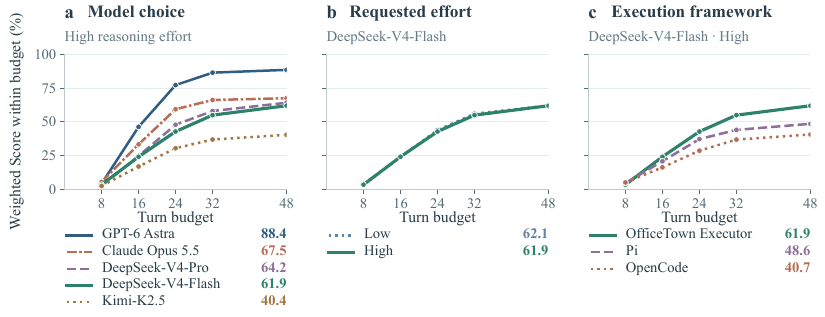}
\caption{OfficeTown Bench at five recorded turn budgets.
All panels use the same \OTBenchmarkTasks{} tasks. Scores enter only after execution
ends; terminal evaluation errors remain zero.
(a) High-effort models with the same executor. (b) Low and High
effort for DeepSeek-V4-Flash. (c) Execution frameworks for DeepSeek-V4-Flash High.
Points show 8, 16, 24, 32, and 48 turns; lines guide the eye.
Appendix~\ref{app:budget-studies} specifies counting and execution settings.}
\label{fig:bench-turn-limits}
\end{figure}

\section{Experiments}
\label{sec:experiments}

We evaluate organizational simulation through the work it produces: whether
OfficeTown tasks challenge current agents, and whether learning from this work
improves external-benchmark performance. A supplementary study tests the
principal--assistant design on external tasks.

\subsection{Evaluation on OfficeTown Bench}
\label{sec:bench-results}

Agents execute work requests using the supplied files and tools, without access
to the task-specific rubrics. We assess submitted artifacts and observable
outcomes through executable checks and content-level judgment.
Table~\ref{tab:otbench-fill} reports Pass@1 for 13 models on the same
\OTBenchmarkTasks{} tasks: a task passes only when all applicable required
criteria are met. Weighted Score, reported in Appendix~\ref{app:weighted-scores},
averages each task's fulfilled rubric-weight fraction. These metrics distinguish
complete delivery from partial fulfillment; scoring and execution protocols are
in Appendix~\ref{app:protocols}.

GPT-6 Astra leads with 74.4\% Pass@1, followed by Grok-4.7 at 58.4\% and
Claude Opus 5.5 at 50.7\%.
Even Astra leaves roughly one quarter of tasks incompletely delivered.
The metrics also expose different performance profiles: Claude passes more
tasks than GPT-5.6 Terra (50.7\% versus 39.4\%) despite a lower Weighted
Score (67.5 versus 76.1). High average fulfillment therefore need not imply
reliable complete delivery. Domain strengths vary: Grok-4.7 exceeds Astra
in assurance (70.2\% versus 63.8\%) despite its lower overall pass rate.

This gap extends to critical requirements (Figure~\ref{fig:bench-diagnostics}).
Astra satisfies 88.8\% of critical criterion weight on average, yet 84 of
404 eligible tasks (20.8\%) miss at least one critical (P0) requirement.
The miss rate rises to 53.3\% for Terra and 78.3\% for DeepSeek-V4-Flash,
while Claude Opus 5.5 and Grok-4.7 miss critical requirements on 41.4\% and
33.3\% of eligible tasks. Failures are not limited to optional refinements. Input-file groups
reveal different relative strengths: Astra leads Terra on CAD-containing tasks,
but trails on media-containing tasks. These overlapping task populations do not
isolate the effect of file format.

Figure~\ref{fig:bench-turn-limits} summarizes five budgets on the same 406 tasks.
In the full records, Astra reaches a Weighted Score of 40 within 16 turns,
followed by Claude at 18, DeepSeek-V4-Pro at 22, DeepSeek-V4-Flash at 24,
and Kimi-K2.5 at 42. Flash's Low and High settings finish close in Weighted
Score (62.1 versus 61.9), whereas execution frameworks differ more: OfficeTown
Executor reaches 61.9, Pi 48.6, and OpenCode 40.7. These are recorded runs under each framework's native prompts and tools, not
token-matched comparisons. OfficeTown Executor is single-agent, distinct from the
principal--assistant harness below.

\subsection{Learning from OfficeTown}
\label{sec:training-results}

\begin{table}[!t]
\centering
\caption{Learning from OfficeTown: Qwen3.5-9B across training stages (\%).
Rubric and Weighted denote mean rubric and weighted rubric scores;
Success denotes the native binary success rate.
Subscripts show percentage-point differences from Base.
Protocols: Appendix~\ref{app:training-evaluation}.}
\label{tab:training-fill}
\begingroup
\OTResultSetup
\begin{tabularx}{\linewidth}{l*{5}{>{\raggedleft\arraybackslash}X}}
\toprule
\textbf{Training} & GDPval &
JobBench-Easy & APEX-Agents &
Terminal-Bench 2.1 & OfficeTown Bench \\
& Rubric & Weighted & Pass@1 & Success & Pass@1 \\
\midrule
\OTStage{OTBase}{Base} & 65.75 &
40.41 &
9.40 &
42.31 &
1.48 \\
\OTStage{OTSFT}{SFT} & 70.30\textsubscript{\textcolor{OTSFT}{+4.55}} &
44.87\textsubscript{\textcolor{OTSFT}{+4.46}} &
16.16\textsubscript{\textcolor{OTSFT}{+6.76}} &
51.22\textsubscript{\textcolor{OTSFT}{+8.91}} &
3.45\textsubscript{\textcolor{OTSFT}{+1.97}} \\
\OTStage{OTOPD}{SFT + OPD} & \textbf{73.65}\textsubscript{\textcolor{OTOPD}{+7.90}} &
\textbf{51.48}\textsubscript{\textcolor{OTOPD}{+11.07}} &
\textbf{19.43}\textsubscript{\textcolor{OTOPD}{+10.03}} &
\textbf{63.64}\textsubscript{\textcolor{OTOPD}{+21.33}} &
\textbf{4.68}\textsubscript{\textcolor{OTOPD}{+3.20}} \\
\bottomrule
\end{tabularx}
\endgroup

\end{table}

To test learning value beyond the source setting, we fine-tune Qwen3.5-9B on
DeepSeek-V4-Flash demonstrations, then continue with OPD~\citep{agarwal2024policy} in OfficeTown
environments using Qwen3.8-Flash-Next as teacher. Table~\ref{tab:training-fill}
compares Base, SFT, and SFT + OPD. External benchmarks use their respective
metrics; GDPval and Terminal-Bench exclude search-dependent tasks
(Appendix~\ref{app:training-evaluation}).

On 79 shared GDPval tasks, the rubric score rises from 65.75 to 70.30 after SFT
and to 73.65 after OPD, gaining 7.90 percentage points over Base.
JobBench-Easy~\citep{li2026jobbench} and APEX-Agents~\citep{vidgen2026apexagents}
also improve across the reported stages.
Together, the results support the learning value of OfficeTown-generated work
beyond its source
organizations. The corresponding OfficeTown Bench Weighted Scores are 26.20,
29.20, and 31.33 for Base, SFT, and SFT + OPD, respectively.
Appendices~\ref{app:training-dynamics}--\ref{app:training-curves} report
training dynamics and retrospective budget curves for these runs.

\subsection{Harness Evaluation}
\label{sec:harness-results}

\begin{table}[!t]

\caption{Harness results on external benchmarks.
All harnesses use GPT-5.5 with xhigh reasoning effort.
Reported point estimates on a 0--100 scale; higher is better.
GDPevo aggregates repeated runs by task group, not best-of-three success.
Codex on GDPevo uses the few-shot reference setting.
Terminal results use different reported budgets.}
\label{tab:harness-fill}

\begingroup
\OTResultSetup
\begin{center}
\begin{tabularx}{\linewidth}{p{0.25\linewidth}*{5}{>{\raggedleft\arraybackslash}X}}
\toprule
\textbf{Harness} & \multicolumn{2}{c}{PRDBench} & GDPevo &
Terminal-Bench 2.1 & Research-code \\
\cmidrule(lr){2-3}
& DEV & DEBUG & & & \\
\midrule
Codex & 85.87 & 89.04 & 64.51 & 83.7 & 40.88 \\
MetaGPT & 84.86 & 86.51 & 63.44 & 81.1 & 41.15 \\
CAMEL & 76.74 & 78.93 & 63.45 & 85.1 & 39.95 \\
\midrule
\rowcolor{OTTableFocus}
OfficeTown Harness & 89.67 & 92.62 & 74.70 & 87.4 & 46.40 \\
\bottomrule
\end{tabularx}
\end{center}
\endgroup

\end{table}

This supplementary study tests whether the principal--assistant design for
delegation, clarification, and review is useful beyond the simulated
organizations. We compare OfficeTown
Harness with Codex, MetaGPT~\citep{hong2024metagpt}, and CAMEL~\citep{li2023camel}
using GPT-5.5 xhigh throughout on external tasks
(Table~\ref{tab:harness-fill}). OfficeTown has the highest reported point
estimate in all five columns, covering initial development and feedback-guided
revision (PRDBench DEV/DEBUG; \citealp{fu2025prdbench}), business rules
(GDPevo; \citealp{zhou2026gdpevo}), terminal work~\citep{merrill2026terminalbench},
and research code~\citep{hua2025researchcodebench}.
Token use varies by benchmark. On research code, OfficeTown scores 46.40
versus MetaGPT's 41.15 with 21.8\% fewer reported tokens, while using 3.49 times
Codex's tokens. Terminal results use different budgets, and Codex's GDPevo
reference includes few-shot skills.

\section{Related Work}
\label{sec:related-work}

\paragraph{Professional Work Benchmarks}
\label{sec:related-environments}

Executable benchmarks cover web and desktop interaction~\citep{zhou2024webarena,xie2024osworld},
enterprise knowledge work~\citep{drouin2024workarena,boisvert2024workarenaplus},
and cross-application, long-horizon, and file-dependent
workflows~\citep{wang2024officebench,wang2025odysseybench,tang2026workspacebench}.
GDPval evaluates professional deliverables across occupations~\citep{patwardhan2025gdpval}.
EnterpriseBench, CRMArena-Pro, and AutomationBench examine cross-functional workflows,
access constraints, and policy-governed interaction~\citep{vishwakarma2025enterprisebench,huang2025crmarena,shepard2026automationbench}.
ClawMark includes multi-day workflows with externally changing service
state~\citep{meng2026clawmark}. Closest to our setting, TheAgentCompany combines a
simulated company, office applications, and model-driven colleagues with manually
curated tasks~\citep{xu2024theagentcompany}.

\paragraph{Environment and Task Synthesis}
\label{sec:related-synthesis}

Environment synthesis scales the construction of executable tasks and training
resources. ScaleEnv generates tool environments and rollout-verified tasks, while
EnvFactory builds stateful environments and training trajectories from API
resources~\citep{tu2026scaleenv,xu2026envfactory}.
EnvScaler, AWM, and AutoForge automate environment and verifiable-task
construction~\citep{song2026envscaler,wang2026awm,cai2025autoforge}.
Agent-World combines synthesis with iterative reinforcement
learning~\citep{dong2026agentworld}, and ClawEnvKit generates tasks, tools, and
scoring criteria from natural-language requests~\citep{li2026clawenvkit}.
ScienceBuddy derives executable tasks and rubrics from researcher
collaboration~\citep{xue2026sciencebuddy}. Our focus is work generation, not
recursive self-improvement~\citep{duan2026rsi}.

\paragraph{Generative Agents and Organizational Simulation}
\label{sec:related-simulation}

Generative Agents uses memory, reflection, and planning to sustain social
activity~\citep{park2023generativeagents}. SOTOPIA studies goal-directed social
interaction~\citep{zhou2024sotopia}, while $\tau$-bench evaluates policy-constrained
tool use with simulated users~\citep{yao2024taubench}.
Agent Hospital connects institutional simulation with professional capability
development through accumulated medical cases and reflection, without updating
the underlying models~\citep{li2024agenthospital}.

OfficeTown connects organizational simulation with executable environment generation.
Earlier artifacts, decisions, and unresolved matters shape subsequent work, while
roles, ownership, and sharing rules scope the information and authority available
for each assignment. This persistent, permissioned process generates tasks,
context, and feedback for evaluation and training.

\section{Conclusion and Future Work}
\label{sec:conclusion}

OfficeTown models professional organizations as persistent, permissioned, and
evolving work systems. Tasks, contextual materials, and feedback arise from
ongoing work, rather than being constructed as isolated instances. Benchmark
evaluations reveal a gap between partial fulfillment and complete professional
delivery, while training results demonstrate the learning value of the generated
work on external benchmarks.

Looking ahead, we will extend these simulations to more professional domains
and institutional settings, with richer interactions, longer-running workflows,
and collaboration across organizations. We will also investigate how persistent
history, scoped access, and review shape task generation and learning outcomes.
Improving simulation fidelity and efficiency, and validating simulated work
against real professional practice, will be important as these systems expand.

\paragraph{Ethics Statement.}
OfficeTown uses referenced public sources, adapted materials, and synthetic
records; it does not use private customer records. Domain experts contributed
feedback on tasks and rubrics, and we report their contributions in aggregate.
Occupational taxonomies and generative models can encode social and professional
biases, so simulated organizations are examples rather than authoritative
representations of occupations, institutions, or demographic groups. Tool-using
agents can create privacy, security, and operational risks. In OfficeTown, agents
operate within permission-scoped workspaces and synthetic authority; they are not
used for real-world financial, legal, clinical, employment, or external-release
decisions. Releases should preserve source attribution and licensing, remove
sensitive operational information, and document intended use and limitations. 
OfficeTown is the name of this research project, not a Microsoft Office product.

\label{paper-main-end}
\bibliography{references}
\bibliographystyle{officetown_arxiv}

\clearpage
\appendix
\addtocontents{toc}{\protect\setcounter{tocdepth}{3}}
\phantomsection
\pdfbookmark[0]{Appendix Contents}{appendix-contents}
\label{app:contents}
\begingroup
\hypersetup{linktoc=all}
\renewcommand{\contentsname}{Appendix Contents}
\tableofcontents
\endgroup
\clearpage
\section{Organization Construction and Simulation}
\label{app:simulation}

\subsection{Creating an Organization from a Natural-Language Description}
\label{app:organization-construction}

\begin{figure}[!htbp]
\centering
\includegraphics[width=\linewidth]{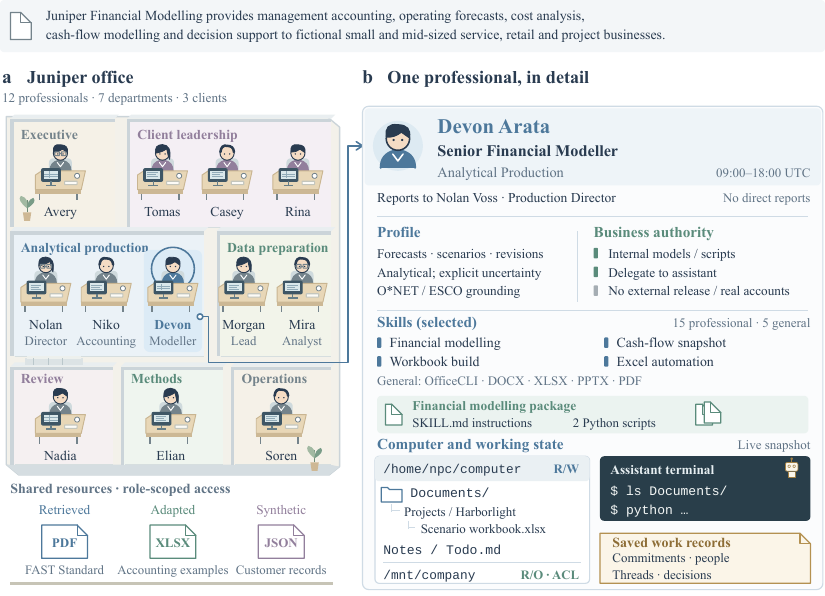}
\caption{A generated office and one professional in detail.
(a) Juniper's 12 professionals, grouped by department, with examples of retrieved,
adapted, and synthetic seed materials. The brief above is an excerpt of the full
request, which also includes configuration.
(b) Devon Arata's role, manager, profile, business authority, selected skills, and
computer. The blue connector opens the highlighted person's details; it is not a
reporting relationship. Profiles and skills describe the generated configuration;
files and saved work records show retained simulation state. The office layout,
avatars, and terminal commands are illustrative. R/W denotes read/write access;
R/O denotes read-only access, filtered by access-control rules (ACL).}
\label{fig:appendix-organization}
\end{figure}

A construction request combines a description of the organization's purpose and
activities (as short as one sentence) with optional settings for organization size,
language, and simulation period. OfficeTown expands the request into a scope of work,
departments, roles, reporting
relationships, and synthetic customers. In the Juniper example
(Figure~\ref{fig:appendix-organization}a), departments divide client engagement,
analytical work, data preparation, and review.

Professional roles are grounded in occupational descriptions from O*NET and ESCO.
Each professional receives a profile specifying their responsibilities, working
style, schedule, and reporting relationships. Figure~\ref{fig:appendix-organization}b
expands one such role: Devon Arata, a Senior Financial Modeller who reports to
Analytical Production Director Nolan Voss. Devon is equipped with 15 professional
and five general skill packages, including financial modelling and spreadsheet
tools. Packages are selected from a shared library, supplemented through skills.sh
when needed, and retain their instructions and supporting resources. The expanded
financial-modelling package contains instructions and two Python scripts.

The role also defines what the professional is authorized to do. Devon can prepare
internal models, run analysis scripts, and delegate work to an assistant, but cannot
release material externally or operate real financial accounts.

To build the seed, OfficeTown uses Web search to locate relevant professional guidance,
reference documents, datasets, and worked examples. Retrieved materials are either
retained in their original form or adapted to the organization's work.
Synthetic records supply the organization-specific information needed to begin
work, such as customer operating data, opening balances, and current business questions.

The seed preserves the distinction between retrieved sources, adapted materials,
and synthetic records, illustrated by the reference PDF, accounting workbook, and
customer data in Figure~\ref{fig:appendix-organization}a. Source references accompany
the materials, while file-format
and structural checks establish that the supplied assets can be read and used by
the corresponding tools. These checks concern the materials' accessibility and
structure; professional judgments about their interpretation remain part of the
simulated work.

OfficeTown assembles the organization description, profiles, skill packages, and
seed into a runnable organization package. This package initializes personal
workspaces, authorized shared resources, and the configured office services.

\subsection{Example Organization and Configuration}
\label{app:example-configuration}

Juniper Financial Modelling (Figure~\ref{fig:officetown-framework}) was constructed
from a brief whose business-specific portion begins:

\begin{quote}
Juniper Financial Modelling provides management accounting, operating forecasts,
cost analysis, cash-flow modelling and decision support to fictional small and
mid-sized service, retail and project businesses.
\end{quote}

The remaining brief specifies the expected work products, review requirements, and
use of synthetic customer records. Table~\ref{tab:simulation-configuration} separates
the generated organization properties from the settings used to construct and run
the instance.

\begin{table}[!htbp]
\caption{Generated properties and recorded settings of the Juniper instance.
The simulation period is a configured horizon, not a measure of completed work.
These settings are specific to this instance.}
\label{tab:simulation-configuration}

\begin{center}
\begin{tabularx}{\linewidth}{p{0.39\linewidth}X}
\toprule
Configuration & Juniper instance \\
\midrule
Generated organization & 12 professionals; 7 departments; 11 reporting relationships \\
Initial synthetic customers & 3 \\
\midrule
Staffing target and limit & Target: 12; maximum: 16 \\
Occupational references & O*NET 30.3; ESCO 1.2.1 \\
Recorded generation model & GPT-5.5 \\
Professional and assistant models & GPT-5.5 \\
Assistant execution & Codex harness; worker containers; \texttt{direct\_home} \\
Configured simulation period & 1 January--1 July 2026, 09:00 UTC; 181 days \\
Clock policy & Adaptive business clock; maximum rate: 120 simulated seconds per wall-clock second \\
Execution capacity & Up to 6 concurrent professional turns; 4 assistant workers \\
Assistant timeout & 1,200 seconds \\
Clock checkpoint interval & 5 seconds \\
Service deployment & Isolated deployment (\texttt{isolated\_compose}) \\
Runtime security profile & \texttt{legacy-dev} \\
Run-level monitoring & Record-only \\
Additional semantic-memory backend & Disabled \\
\bottomrule
\end{tabularx}
\end{center}
\end{table}

The seed combines professional references and an adapted management-accounting
workbook with three synthetic customer packets containing 34 groups of business
records. References include the FAST Standard and the ICAEW Financial Modelling
Code. Harborlight's packet contains invoices, payables, payroll records, dispatch
jobs, opening balances, and recent bank activity, accompanied by control totals,
source references, and unresolved assumptions.

The Harborlight packet anchors the 13-week liquidity discussion in
Figure~\ref{fig:appendix-dialogue}, which is described next.

\FloatBarrier
\subsection{Communication and Work Coordination}
\label{app:communication}

OfficeTown connects professionals to running office services through service-backed
tools: Rocket.Chat for direct messages and team channels, Mailpit for email, and
Ba\"ikal for calendars and meeting scheduling. Deliveries are also recorded in persistent,
recipient-specific inboxes, which the simulation uses to present incoming work.
A recipient receives the message and permitted references, not the sender's
private working context.

The meeting organizer specifies the participants, time, and agenda. Each
participant responds through their own model context, with preceding contributions
supplied as part of the discussion. Turn-taking and explicit mentions determine
who speaks next.

Work agreed upon in conversation is recorded through explicit actions. Requests
identify the intended outcome, responsible person, relevant materials, and deadline.
A professional can seek clarification, delegate execution to an assistant, or review
a returned deliverable. Meeting notes summarize the discussion, but do not themselves
create an approval or mark an assignment as complete. These changes require the
corresponding work actions.

\begin{figure}[!htbp]
\centering
\includegraphics[width=\linewidth]{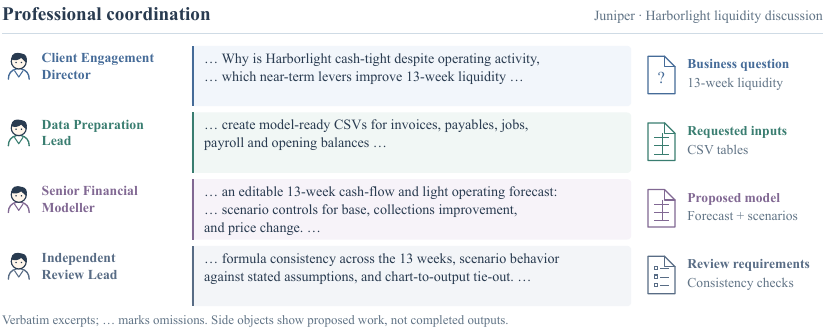}
\caption{Professional coordination in a simulated office. Four consecutive speaking
turns from a Juniper meeting about a synthetic customer's liquidity. Excerpts retain
the original wording and order; ellipses mark omitted text. Side objects identify
the proposed inputs, model, and review requirements, not completed or validated outputs.}
\label{fig:appendix-dialogue}
\end{figure}

Figure~\ref{fig:appendix-dialogue} shows four consecutive turns from a Juniper
meeting about a synthetic customer's liquidity. Client engagement, data preparation,
modelling, and review roles establish the required inputs, handoffs, forecast
scenarios, and checks before execution.

\subsection{Sessions and Memory}
\label{app:sessions-memory}

\begin{figure}[!htb]
\centering
\includegraphics[width=\linewidth]{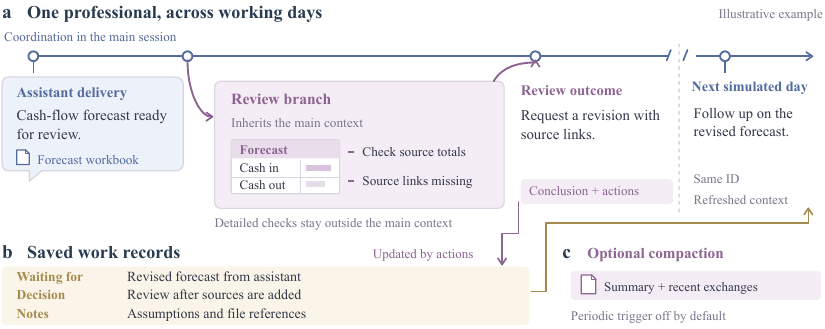}
\caption{Continuing work through sessions and saved records.
(a) In persistent-session mode, the same professional inspects a returned forecast
in a review branch and brings its conclusion, actions, and a brief tool record
back into coordination. (b) Actions update work records; selected records and
personal notes remain available when the context is refreshed under the same
identifier on the next simulated day. (c) Compaction is optional; periodic
triggering is off by default. The forecast, messages, and records form an
illustrative example, not a recorded Juniper interaction.}
\label{fig:appendix-memory}
\end{figure}

Work records are stored independently of active conversation context. They retain
commitments, ongoing discussions, unresolved decisions, and recent interactions
with colleagues (Figure~\ref{fig:appendix-memory}(b)). Commitments record what is
owed and to whom; discussions identify who should respond next. Communication and
work actions update these records through explicit requests and rule-based message
cues. Clearing an incoming reminder does not mark the promised deliverable as
complete. These structured records are distinct from the additional semantic-memory
backend, which is disabled in the Juniper configuration
(Table~\ref{tab:simulation-configuration}).

Only selected records enter each turn. Outstanding promises take priority over
routine reminders, and discussions awaiting the professional's reply take priority
over other exchanges. The default limits are eight active commitments, five
discussions, five unresolved decisions, and ten colleague records. Professionals
can also maintain personal notes containing working assumptions, document references,
and follow-up plans. Recent notes are included within a separate size budget, while
current project and request states are read from the simulation's stored records.

OfficeTown also supports an optional persistent-session mode for continuing
coordination. The main session retains available conversation context alongside
current requests and project state. Detailed inspections take place in a branch
of the same professional's session (Figure~\ref{fig:appendix-memory}(a)). The branch
inherits the current context and supports file inspection and tool use. In the
illustrative forecast review, the professional identifies missing source links
and requests a revision. A concise record of the conclusion, actions, and tools
used returns to the main session, without retaining the full inspection in its
active context.

At simulated day boundaries in this mode, the main context is refreshed by
reopening the session under the same identifier with saved work records and
current office state. The pending forecast revision in the illustration can
therefore inform a later turn without preserving the full earlier conversation.
OfficeTown also supports optional compaction, which summarizes earlier exchanges
while retaining recent messages (Figure~\ref{fig:appendix-memory}(c)). Its periodic
trigger is disabled by default.

\subsection{Workspaces, Tools, and Access Control}
\label{app:workspaces}

\begin{figure}[!htb]
\centering
\includegraphics[width=\linewidth]{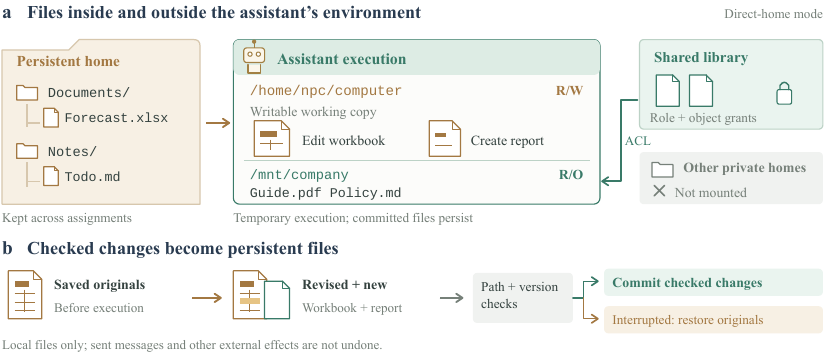}
\caption{A bounded file view and persistent local changes.
(a) In direct-home mode, an assistant receives a writable copy of its principal's
home and a permission-filtered, read-only company view. Other private homes are
not mounted. R/W and R/O denote read/write and read-only access; ACL denotes
access-control rules. (b) The runtime checks local changes before committing them;
an interrupted commit can restore the preceding files and classification state.
The files and changes are illustrative, not execution results. Recovery does not
undo messages or other effects already sent to external services.}
\label{fig:appendix-workspace}
\end{figure}

An assistant works in a temporary execution container, while the principal's
computer workspace persists across assignments (Figure~\ref{fig:appendix-workspace}(a)).
Professionals use this workspace to inspect and edit files, execute terminal
commands, and access their skill packages.

In direct-home mode, the runtime provides a writable copy of the principal's current home at
\texttt{/home/npc/computer}, together with authorized shared references. A runtime-generated
instruction block identifies the principal and distinguishes writable files, read-only
inputs, and temporary scratch space. Tasks requiring narrower access instead receive
an isolated view with selected inputs and designated working and submission directories.
Committed work remains in the professional's persistent workspace after the
execution container is released.

Shared materials are presented through a separate, permission-filtered company view
at \texttt{/mnt/company}. Ownership, role permissions, explicit sharing rules, and
confidentiality classifications determine which documents appear in this view.
A reporting relationship does not grant unrestricted access to another person's files.
Other professionals' private homes are not mounted in the assistant's environment.
Shared documents are read-only: professionals and assistants can consult them or create
their own working copies, but cannot directly overwrite the shared originals. File
operations and terminal execution enforce these boundaries through path checks,
read-only mounts, and sandboxing.

The revised workbook and new report in Figure~\ref{fig:appendix-workspace}(b)
enter the persistent workspace only after the runtime checks protected paths,
file validity, and the underlying workspace revision. Writers to the same
professional's home are serialized to avoid conflicting updates. The runtime records
the actual files created, modified, moved, or deleted, independently of the assistant's
completion report. The professional receives the deliverable locations, reported
checks, and unresolved issues for review. Filesystem validation establishes what was
saved; professional review determines whether the content meets the assignment's
requirements.

For broader publication, the professional selects an accepted deliverable and its
intended audience. The runtime checks the acceptance record, file version,
classification, and recipient permissions before making it available to authorized
colleagues. Personal working materials can therefore accumulate without exposing
every intermediate draft to the whole organization.

\subsection{Simulation Time and Recovery}
\label{app:execution-recovery}

Simulated time is managed separately from wall-clock execution time. When adaptive
pacing is enabled, OfficeTown slows or pauses the simulated clock as unfinished work
accumulates, then resumes its advance as the backlog decreases. The clock is periodically
saved together with the run's original start and target times. A resumed run continues
toward that target rather than beginning another full simulation period.

Routine activations follow working hours and meeting schedules; designated
interruptions can receive earlier attention. Professionals act asynchronously
within limits on concurrent model calls and assistant workers. Work awaiting
capacity remains pending.

Incoming messages remain in persistent inbox files, with acknowledgments tied to
the exact source rows processed. Reading or queuing a message does not acknowledge
it as processed. A partially processed batch retains acknowledgments even when
rows finish out of order, leaving unprocessed rows available for later delivery.
Capacity-deferred delegations retain their original identities and arguments when
reconsidered for execution.

Workspace recovery follows the checked-update boundary shown in
Figure~\ref{fig:appendix-workspace}(b) and described in Section~\ref{app:workspaces}.
The runtime saves the preceding workspace state and journals an update before committing
changes. An interruption before commit leaves the persistent workspace unchanged;
an interruption during commit can be recovered by restoring the saved files and
classification metadata. The recovery is recorded explicitly. This mechanism applies
to local workspace changes, not to reversing an email already sent or another external
service action.

The runtime also monitors its action bridge and model-session service and supports
replacing failed services. When the simulation process remains alive, active sessions
can be reopened using saved conversation records and retained session metadata. This
reconstructs context for subsequent turns rather than resuming an interrupted model
generation internally. Transient connection failures receive bounded retries, and
unsuccessful attempts remain recorded.

Run-level monitoring can stop execution on detected violations or record them while
allowing an endurance run to continue. The latter policy leaves operation-level
authorization and validation checks in force. At shutdown, the runtime stops new
activity, allows a bounded interval for ongoing work to settle, and reconciles assistant
results with delivery records. Unfinished requests, failed attempts, and incomplete
reviews remain in the saved state; reaching the target time does not mark them as finished.

\subsection{A Six-Month Office Simulation}
\label{sec:longitudinal-analysis}

\begin{figure}[!htbp]
\centering
\includegraphics[width=\linewidth]{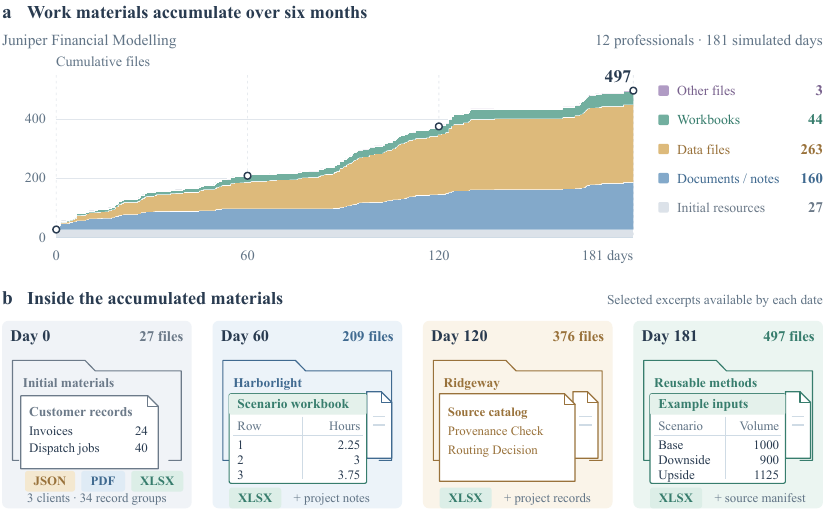}
\caption{Work materials accumulated during a six-month Juniper simulation.
All dates and days denote simulated calendar time, not wall-clock runtime;
the adaptive clock adjusts its rate to work-processing pressure.
(a) Cumulative material files: \JuniperInitialFiles{} distinct initial resource
contents and \JuniperOutputFiles{} logical output files, counted at first
simulated delivery. Revisions and recorded renames count once; exact copies are
merged at first appearance. Navigation READMEs, explicit backups, and trash paths
are excluded. Initial resources are grouped separately from subsequent outputs.
(b) Selected material excerpts available by each date, redrawn from retained
source versions; the final workbook contains synthetic scenario inputs.
Counts describe the whole organization, not the illustrated folders.}
\label{fig:juniper-live}
\end{figure}

We examine material accumulation in the Juniper instance over 181 simulated days,
from 1 January to 1 July 2026. These dates and all checkpoints in
Figure~\ref{fig:juniper-live} follow the simulation clock
(Section~\ref{app:execution-recovery}), not elapsed wall-clock time. Materials are
counted at their first recorded simulated delivery, with repeated revisions counted once.

Figure~\ref{fig:juniper-live}(a) shows the cumulative collection growing from
\JuniperInitialFiles{} initial files to \JuniperDaySixtyFiles{},
\JuniperDayOneTwentyFiles{}, and \JuniperTotalFiles{} files by simulated days 60,
120, and 181, respectively. The additional materials span documents and notes,
data files, workbooks, and other formats. Figure~\ref{fig:juniper-live}(b) places
these counts alongside selected materials available by each checkpoint: customer
operating records, a scenario workbook, a retail-project source catalog, and a
reusable workbook component.
This observation describes material accumulation; it does not assess deliverable
quality or task success.

\FloatBarrier
\section{Data Construction and Benchmark Cases}
\label{app:data}

\subsection{Resource Membership and Separation}
\label{app:data-separation}

Figure~\ref{fig:officetown-data}, Table~\ref{tab:officetown-resources}, and
Tables~\ref{tab:data-composition}--\ref{tab:data-deliverables} describe the same
\OTBenchmarkTasks{} benchmark tasks. Benchmark tasks, SFT demonstrations, and OPD
environments have separate membership records; training-resource construction is
described in Appendix~\ref{app:training}.

Table~\ref{tab:officetown-resources} counts source organizations by their recorded
organization identities, counting repeated simulations of the same identity once.
Professional roles are distinct job titles of the primary delegating professionals,
with case and whitespace normalized; they are not personnel counts or standardized
occupational categories.

Input-file counts include supplied company and personal-workspace files,
historical service readbooks, and supplied service-state records.
They exclude task-packaging metadata, runtime and
evaluation files, and duplicate path aliases; an archive counts as one supplied
file. The task-wise total counts shared materials once per task environment;
unique files are deduplicated across environments by their SHA-256 content hash.
The table reports per-task means and medians.

Rubric counts refer to the \OTRubricCriteria{} criteria in the final evaluation
instrument, not their subordinate verification checks.
Task and criterion selection used multi-model pilot results; eligible pilot judgments
were reused for retained tasks. Aggregate statistics, model comparisons, and case
studies use the same fixed task and criterion sets.

Source task identities remain recorded, but task-identity separation alone does not
establish project-level or semantic isolation.
Exposure to external task instances, answers, rubrics, and test
feedback is distinct from using a benchmark's format or methodology.

\subsection{Expert Review}

To ground tasks and evaluation criteria in professional practice, we engaged
71 domain experts across eight professional domains during benchmark construction.
Task-specific expertise requirements covered relevant specialization and
occupational background, the ability to make independent professional judgments,
and familiarity with native working materials and applicable standards.
The construction-stage annotation records cover 1,051 candidate tasks and
32,274 rubric criteria. Among 1,011 recorded task-level familiarity assessments,
98.9\% indicated that reviewers were familiar or very familiar with the assigned
work. Experts assessed task realism and feasibility and examined rubric
correctness, coverage, and importance, providing 3,842 criterion-level
recommendations to edit, remove, merge, or split criteria. The annotations
subsequently underwent a separate review pass, combining expert assessment with
follow-up quality control and producing refinement feedback traceable to
individual tasks and criteria.

The review instructions ask reviewers to assess task realism,
objective clarity, input sufficiency, feasibility, deliverable clarity, permission
boundaries, answer leakage, and unnecessary restrictions. Rubric review separately
examines task alignment, professional correctness, assessability, clarity,
redundancy, completeness, importance, evidence gaps, and valid alternative solutions.

Reviewers record separate overall decisions for the task and rubric, with reasons
and proposed revisions. For individual criteria, they can retain the standard or
recommend the revisions described above, identify missing requirements, and assign
the importance levels introduced in Section~\ref{sec:officetown-bench}. These decisions
concern the evaluation instrument, not an agent's answer. Expertise or evidence
limitations are recorded rather than treated as approval.

Critical (P0) failures make the work essentially invalid or unusable; Major (P1)
failures substantially affect its intended outcome or professional usability. Minor
(P2) deficiencies do not substantially undermine the core outcome; Optional (P3)
criteria describe desirable refinements with limited impact on overall success.

Concrete edits remain linked to the reviewed task and criterion identities.
For the final benchmark, historical expert feedback is bound to 394 of the 406
retained tasks; the other 12 have model-based review without a bound human annotation.
Subsequent task and rubric adaptation and criterion selection do not constitute
a new round of human review.

\subsection{Domain and File-Type Coverage}

Table~\ref{tab:data-composition} expands the domain overview in
Figure~\ref{fig:officetown-data}a with exact task and source-role totals.
The eight domain petals cover all \OTBenchmarkTasks{} tasks, with equal angles
and areas proportional to task counts. Domain-level title counts are not
additive because the same title can occur in more than one domain.

\begin{table}[!htbp]

\caption{OfficeTown Bench domains and source role titles.
Role titles are deduplicated within each domain; titles can recur across domains. The total uses global deduplication.}
\label{tab:data-composition}

\begin{center}
\begin{tabular}{lrr}
\toprule
Professional domain & Tasks & Role titles \\
\midrule
Assurance & 47 & 14 \\
Clinical & 28 & 14 \\
Design & 103 & 25 \\
Entertainment & 55 & 17 \\
Finance & 67 & 26 \\
Industrial & 48 & 18 \\
Legal & 13 & 10 \\
Software & 45 & 18 \\
Total (distinct titles) & 406 & 139 \\
\bottomrule
\end{tabular}
\end{center}

\end{table}

\begin{figure}[!tp]
\centering
\includegraphics[width=\linewidth]{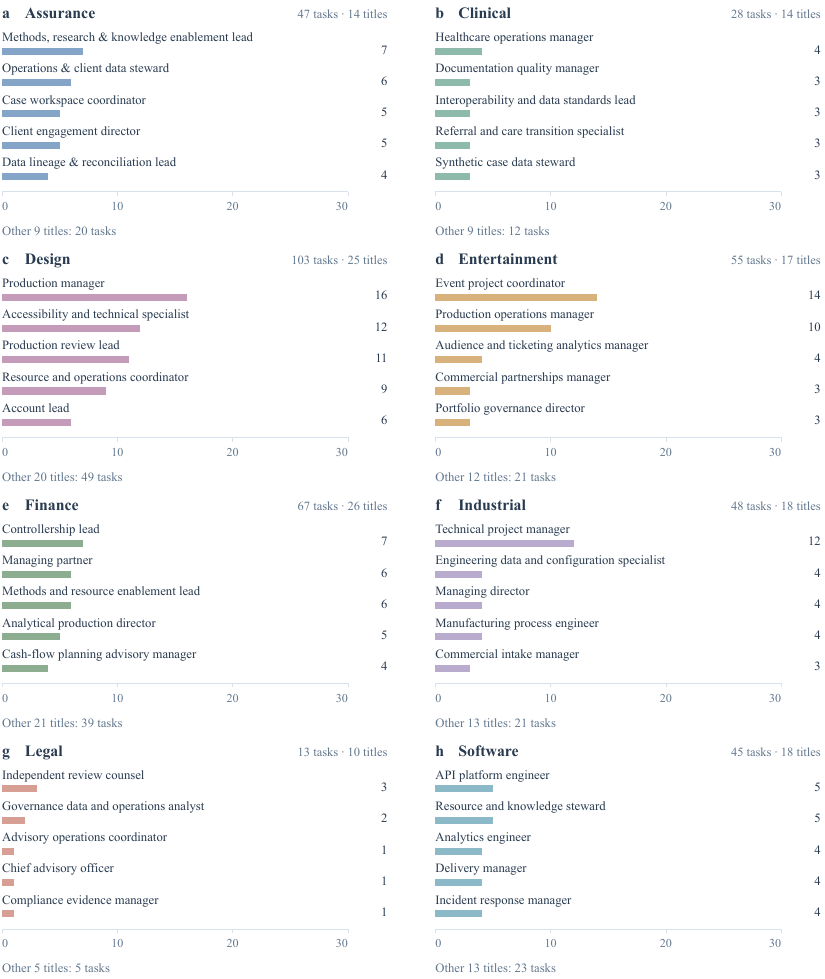}
\caption{Source roles in OfficeTown Bench. Each panel shows the five most frequent
delegating role titles and their task counts on a shared scale. Remaining titles
and tasks are summarized below each panel. Titles are source job names, not
standardized occupations.}
\label{fig:officetown-roles}
\end{figure}

Figure~\ref{fig:officetown-roles} resolves each domain into the five source role
titles associated with the most tasks, with remaining titles and tasks summarized
below each panel. Ties are ordered alphabetically.

Table~\ref{tab:data-deliverables} gives exact input-type counts across all
\OTBenchmarkTasks{} tasks, complementing Figure~\ref{fig:officetown-data}b.
Input coverage counts a file family once per task, including actual members of
supplied ZIP and other archives. Archives are inspected recursively without
unpacking Office documents into their internal XML components. Input coverage
describes available materials, including reference repositories and dependencies;
it does not assert that every file is necessary or used by an agent. File counts
and type coverage use different units: an archive is one supplied file, but can
contribute several types.

Categories overlap. Text and markup include TXT, Markdown, and HTML;
documents include PDF and Word;
tables include spreadsheet and CSV formats; structured data includes JSON, XML,
and configuration files. Code, images, CAD/3D, and audio/video are
classified by their file formats. Unclassified inputs remain in Other files.
These are file-carrier categories, not occupational or task-purpose labels.
These counts describe supplied inputs, not model-generated outputs.

\begin{table}[!htbp]

\caption{Input-file type coverage in OfficeTown Bench.
Each type is counted once per task across all 406 tasks.
Input types include archive contents. Categories overlap.}
\label{tab:data-deliverables}
\begin{center}
\begin{tabular}{lr}
\toprule
File type & Input tasks \\
\midrule
Text / markup & 406 \\
Documents & 318 \\
Tables & 280 \\
Structured data & 406 \\
Code & 250 \\
Images & 195 \\
CAD / 3D & 33 \\
Audio / video & 55 \\
Other files & 126 \\
\bottomrule
\end{tabular}
\end{center}

\end{table}

Figure~\ref{fig:officetown-data}c uses equal-width bins of 20 input files over the
full observed range, starting at the observed minimum of 11 files. The maximum
is 390 files; the last bin ends at 391. Bar height is the number of tasks in each bin. The curve
is a Gaussian kernel density estimate with a 10-file bandwidth, reflected at
zero and multiplied by the task count and bin width to match the frequency axis.
Only the displayed range is drawn; the observed minimum is not treated as a
population support boundary.
It summarizes the same observations, not uncertainty or additional samples.
All \OTBenchmarkTasks{} tasks are included; no tail is truncated.

\FloatBarrier
\clearpage
\subsection{Benchmark Case Studies}
\label{app:benchmark-cases}

The cases in Figures~\ref{fig:case-riverbend} and~\ref{fig:case-oakridge}
illustrate different professional requirements: tracing financial discrepancies
to accounting evidence, and relating fixture geometry to calculations and
inspection access. Complete requests, criteria, exact weights, and initial-file
inventories are retained in the case records.

\subsubsection{Financial Reconciliation}
\label{app:case-riverbend}

In the Riverbend case, a Reconciliation Operations Manager requests Excel
workpapers for three exception streams: customer deposits, card clearing, and
vendor credits. The agent receives an input brief, an operating-data packet,
a native workbook, and a triage memo. It must reconcile the available records,
explain unresolved differences, and identify the evidence needed before an
adjustment or period-close conclusion can be supported.

\begin{figure}[!htb]
\centering
\includegraphics[width=\linewidth]{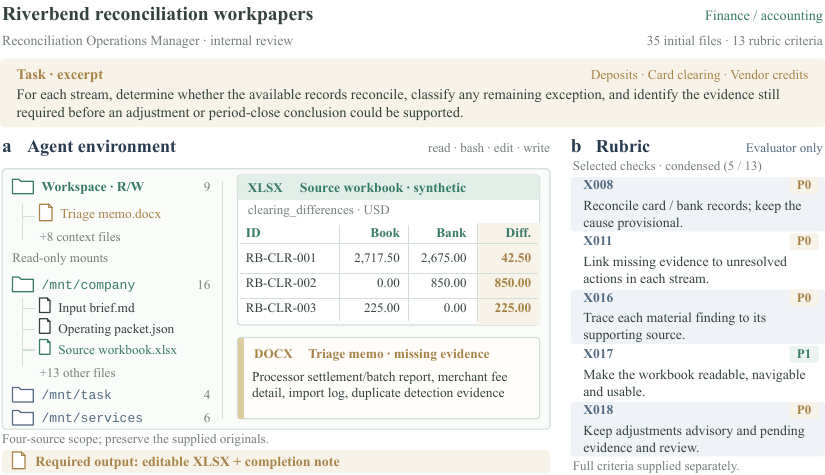}
\caption{Riverbend reconciliation case.
(a) Initial files, selected columns from the supplied workbook, and a
missing-evidence entry from the triage memo. Amounts are synthetic task inputs,
not model results. (b) Five condensed checks from the 13-criterion evaluator-only
rubric. P0 and P1 denote critical and major importance. Filenames and table
headings are shortened; the 35 initial files include task and metadata carriers.
Service files are frozen historical projections. Required outputs appear below
the input tree; the supplied originals must remain unchanged.}
\label{fig:case-riverbend}
\end{figure}

The card-clearing entry in Figure~\ref{fig:case-riverbend}(a) contains a \$42.50 difference;
the memo identifies missing settlement records and fee details needed to
investigate it. The requested workpapers must reconcile the three listed
differences to \$1,117.50, cite the supporting records, and keep unsupported
conclusions open. The 13-criterion rubric assesses numerical consistency, the
basis for each proposed disposition, and editable-workbook usability (selected
checks in Figure~\ref{fig:case-riverbend}(b)). A balanced total alone does not
establish that an exception is resolved.

\FloatBarrier
\clearpage
\subsubsection{Mechanical Fixture Design}
\label{app:case-oakridge}

In the Oakridge case, a Mechanical Design Manager requests two materially
distinct clamp-and-locate fixture concepts for a supplied workpiece.
The starting materials include a STEP model, datum and feature records,
operating assumptions, and engineering references. The agent must preserve the
workpiece and provide editable geometry or reproducible geometry-generator
source for each concept, together with calculations, annotated SVG comparison
views, a technical note, and a package manifest.

\begin{figure}[!htb]
\centering
\includegraphics[width=\linewidth]{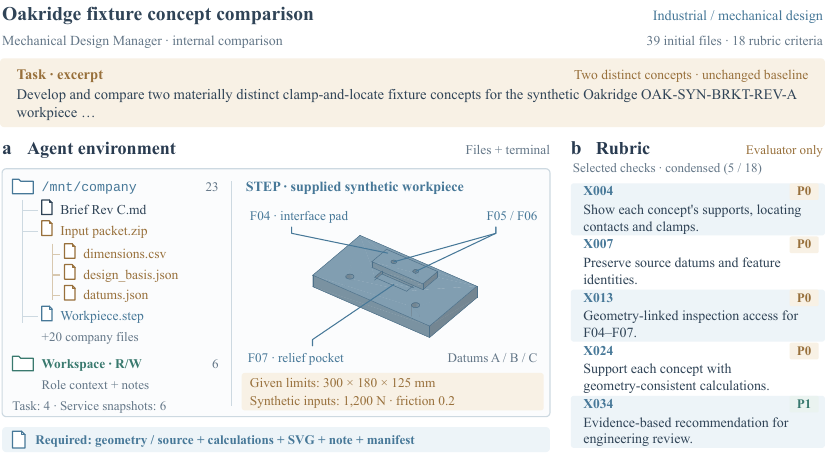}
\caption{Oakridge fixture-design case.
(a) Initial files and the supplied synthetic STEP workpiece, not a proposed
fixture. Indented ZIP children are archive members; filenames are shortened.
The 39 initial files include task and metadata carriers. The given limits
specify the maximum fixture footprint and stack height, not workpiece dimensions.
The load and friction coefficient are synthetic inputs. (b) Five condensed
checks from the 18-criterion evaluator-only rubric, with the same importance
notation as Figure~\ref{fig:case-riverbend}. Required deliverables appear below
the input tree.}
\label{fig:case-oakridge}
\end{figure}

Relevant workpiece features include the interface pad, pin holes, and relief
pocket (Figure~\ref{fig:case-oakridge}(a)).
The design must respect the given equipment envelope and permit the required
inspection access; using the pin holes as functional locators remains
unresolved in the supplied evidence. The 18-criterion rubric checks whether
the geometry can be inspected or rebuilt, whether access claims agree with
the geometry, and whether the supporting calculations can be reproduced
(selected checks in Figure~\ref{fig:case-oakridge}(b)).
Rendered views alone are insufficient, and the recommendation
is for further engineering review rather than production approval.

\FloatBarrier
\label{app:benchmark-cases-end}
\clearpage
\section{OfficeTown Bench Evaluation}
\label{app:experimental-details}
\label{app:protocols}

\subsection{Metrics and Scoring}
\label{app:data-quality}

OfficeTown Bench evaluates submitted artifacts and observable outcomes against
its fixed set of 4,836 task-specific criteria. Weighted Score first measures
fulfilled rubric weight within each task, then averages across tasks.
Let $G_i(c)$ be criterion $c$'s original weight group before filtering, and
$p(c)$ its importance tier. The criterion weight is
\[
w_{ic}=\frac{\alpha_{p(c)}}{|G_i(c)|},
\qquad (\alpha_{P0},\alpha_{P1},\alpha_{P2},\alpha_{P3})=(25,13,7,3).
\]
These are proportional to rank-order centroid weights for the four tiers~\citep{roszkowska2013rank}.
Filtering retains the original shares: removing a criterion does not redistribute
its weight to surviving members. For the retained applicable criteria $A_i$,
\[
s_i=\frac{\sum_{c\in A_i}w_{ic}\,\mathbf{1}[y_{ic}=\mathrm{met}]}
{\sum_{c\in A_i}w_{ic}},
\qquad \mathrm{WS}=\frac{100}{N}\sum_{i=1}^{N}s_i.
\]
Here $N$ is the number of tasks in the overall or domain population.
Non-applicable criteria are excluded; unmet and unassessable criteria contribute
zero. Tasks without applicable criteria score zero and do not pass. Each task
has equal weight in the average; criteria are not pooled across tasks.
Pass@1 requires at least one applicable required criterion and first-attempt
satisfaction of every such criterion.

GPT-5.5 high judges Table~\ref{tab:otbench-fill}; reused judgments require matching
tasks, answers and criteria. Designated terminal evaluation failures score zero,
with failure labels retained: Claude Opus 5.5 and Grok-4.7 include 14 and 15 such
outcomes, respectively. These task-level zeros do not create criterion judgments
or apply to pending evaluations. Sol's Overall Weighted Score remains pending.
Expert review of tasks and rubrics (Appendix~\ref{app:data}) is separate from
model-output scoring and does not establish automated-judge agreement with human
graders. Release and scoring bindings accompany the table.

\subsection{Reasoning and Interaction Budgets}
\label{app:budget-studies}

A logical turn is one model-response round and may contain several tool calls.
Claude's ordered content fragments within one response count as one turn;
native Pi and OpenCode records distinguish admitted calls from streamed updates
and local budget-rejection messages. HTTP retries are not additional logical
turns. Each framework retains its native prompts and tools.

The curves use recorded final grades and execution-ending turns. At budget $T$,
an outcome contributes its final score if execution has ended within $T$ turns,
and zero otherwise, with a fixed denominator. This includes graded partial
outcomes from executions without a final answer. Nonnegative score accumulation
makes the curves monotone; it neither scores intermediate artifacts nor reruns
tasks under shorter stopping budgets.
The figures display five fixed budgets (8, 16, 24, 32, and 48 turns), with
connecting lines as visual guides; exact threshold crossings use the full records.

All three panels use the same \OTBenchmarkTasks{} tasks and final rubric.
The model panel uses the verl-based single-agent OfficeTown Executor, distinct
from the principal--assistant OfficeTown Harness. Shared limits are 48 logical
turns, 900 seconds and 262,144 total output tokens, with a 32,768-token ceiling
per request; Claude Opus 5.5 retains its 16,000-token request cap.

The effort panel compares requested Low and High for DeepSeek-V4-Flash with the
same inputs, environments, executor and shared limits; Medium is not included.
The framework panel fixes DeepSeek-V4-Flash at High. Terminal evaluation errors
remain task-level zeros: 34 for Low, 4 for High in OfficeTown Executor, 18 for Pi,
and 8 for OpenCode. Ordinary judged failures also remain included; tasks are not
dropped to equalize valid-judgment counts. These zeros follow
Appendix~\ref{app:data-quality}, without invented criterion judgments.
Figure~\ref{fig:bench-turn-limits-pass} uses the same records with binary passes.

Training evaluation protocols and curves are in
Appendices~\ref{app:training-evaluation} and~\ref{app:training-curves}.
Both curve analyses reuse existing executions and grades, without new inference,
intermediate-state replay or grading.

\subsection{Coverage and Resource Accounting}
\label{app:resources}

Coverage distinguishes scored attempts, agent errors, budget exhaustion,
infrastructure failures and grading failures. Unreported scores remain unresolved;
the designated task-level zeros follow Appendix~\ref{app:data-quality}.
Shared-task comparisons use paired observations. Repeated attempts and criteria
within a task are not independent samples, and tasks from the same organization
or project may also be dependent; aggregate means alone do not determine
confidence intervals.

For multi-role systems, accounting must distinguish principal, assistant, reviewer
and retry activity, reporting native outer-loop counts separately from response
rounds summed over roles. Input, cached-input, output and available reasoning
tokens retain their original meanings and are not double-counted. Neither equal
turn limits nor completion thresholds establish equal token or monetary cost;
monetary comparisons also depend on the pricing version.

\section{OfficeTown Bench Analysis}
\label{app:diagnostics}

\subsection{Weighted Scores by Domain}
\label{app:weighted-scores}

Table~\ref{tab:otbench-weighted} complements the Pass@1 leaderboard with
rubric-weighted performance on the same tasks. It captures partial fulfillment
under the weighting, applicability and missing-outcome rules in
Appendix~\ref{app:data-quality}.

\begin{table}[H]
\caption{OfficeTown Bench Weighted Score (\%; higher is better).
\colorbox{OTRankOne}{1st} \colorbox{OTRankTwo}{2nd}
\colorbox{OTRankThree}{3rd}: ranks within each column at displayed precision;
ties share rank and skip positions. Models, tasks and high-effort settings match
Table~\ref{tab:otbench-fill}. Dashes indicate pending results. Domain abbreviations denote assurance, clinical, design,
entertainment, finance, industrial, legal, and software work.}
\label{tab:otbench-weighted}
\begingroup
\begin{center}
\begin{tabularx}{\linewidth}{Yrrrrrrrrr}
\toprule
\textbf{Model} & Overall & Assur. & Clin. & Des. & Ent. & Fin. & Ind. & Legal & Soft. \\
\midrule
\rowcolor{OTTableGroup}\multicolumn{10}{l}{\textit{OpenAI}} \\
GPT-6 Astra & \OTRank{OTRankOne}{88.4} & \OTRank{OTRankOne}{85.8} & \OTRank{OTRankOne}{97.9} & \OTRank{OTRankTwo}{79.3} & \OTRank{OTRankOne}{85.3} & \OTRank{OTRankOne}{93.7} & \OTRank{OTRankOne}{91.1} & \OTRank{OTRankOne}{92.3} & \OTRank{OTRankOne}{98.2} \\
GPT-5.6 Sol & 68.5 & 67.7 & 64.2 & 66.7 & 72.4 & 68.1 & 73.2 & \OTRank{OTRankThree}{81.7} & 62.7 \\
GPT-5.6 Terra & \OTRank{OTRankThree}{76.1} & 80.0 & 64.8 & \OTRank{OTRankOne}{79.7} & 78.0 & 73.5 & \OTRank{OTRankTwo}{75.3} & 70.6 & \OTRank{OTRankThree}{75.0} \\
GPT-5.6 Luna & 75.4 & \OTRank{OTRankTwo}{83.6} & \OTRank{OTRankTwo}{84.1} & 71.6 & \OTRank{OTRankTwo}{81.4} & \OTRank{OTRankThree}{74.0} & 71.6 & \OTRank{OTRankTwo}{88.4} & 64.9 \\
GPT-5.4 & 72.2 & 73.0 & 75.7 & 70.0 & \OTRank{OTRankThree}{80.5} & 69.8 & 70.9 & 59.7 & 72.2 \\
GPT-5.2 & 64.3 & 65.4 & 71.5 & 59.4 & 73.1 & 63.9 & 62.0 & 61.2 & 63.1 \\
\rowcolor{OTTableGroup}\multicolumn{10}{l}{\textit{Anthropic}} \\
Claude Opus 5.5 & 67.5 & 68.8 & 78.0 & 63.1 & 71.0 & 70.4 & 61.8 & 75.6 & 64.6 \\
\rowcolor{OTTableGroup}\multicolumn{10}{l}{\textit{DeepSeek}} \\
V4-Pro & 64.2 & 66.4 & 67.9 & 62.3 & 71.8 & 61.0 & 62.1 & 54.9 & 63.9 \\
V4-Flash & 61.9 & 70.1 & 62.7 & 58.8 & 63.5 & 59.9 & 56.7 & 67.1 & 65.2 \\
\rowcolor{OTTableGroup}\multicolumn{10}{l}{\textit{Moonshot AI}} \\
Kimi-K2.5 & 40.4 & 39.0 & 43.6 & 41.3 & 45.1 & 35.9 & 35.4 & 46.1 & 42.9 \\
\rowcolor{OTTableGroup}\multicolumn{10}{l}{\textit{xAI}} \\
Grok-4.7 & \OTRank{OTRankTwo}{77.2} & \OTRank{OTRankThree}{81.9} & \OTRank{OTRankThree}{83.3} & \OTRank{OTRankThree}{74.1} & 76.1 & \OTRank{OTRankTwo}{79.2} & \OTRank{OTRankThree}{75.0} & 73.8 & \OTRank{OTRankTwo}{77.3} \\
Grok-4.3 & 43.3 & 54.4 & 47.0 & 41.6 & 48.9 & 37.4 & 34.6 & 35.8 & 46.3 \\
\rowcolor{OTTableGroup}\multicolumn{10}{l}{\textit{Mistral AI}} \\
Mistral-Large-3 & 20.3 & 25.0 & 17.3 & 19.7 & 29.7 & 11.2 & 18.6 & 18.2 & 23.3 \\
\bottomrule
\end{tabularx}
\end{center}
\endgroup

\end{table}

\subsection{Requirement Importance}
\label{app:requirements}
\label{app:artifact-analysis}

Figure~\ref{fig:bench-diagnostics} and Table~\ref{tab:importance-scores} use the
retained P0--P3 labels for critical, major, minor and optional requirements,
preserving their expert-inherited, expert-adapted or model-proposed provenance.
The twelve models supply 4,825 valid task-level judgments from 4,872 assigned
executions. The other 47 outcomes retain their declared task-level zeros without
invented criterion failures. Models lacking bound item-level records are omitted,
not replaced with partial or historical evaluations.

Within each task and tier, fulfillment is the ratio of satisfied to applicable
criterion weight, averaged across eligible tasks. A critical miss occurs when at least one applicable P0 criterion is unmet
or unassessable. Its rate uses validly judged, P0-bearing tasks; Pass@1 instead
covers all required criteria under the main table's reporting population.

\begin{table}[H]
\caption{Requirement fulfillment by importance, and tasks with a critical miss (\%).
Fulfillment is task-averaged; higher is better. P0 misses are task-level
rates; lower is better. Eligible task counts vary by model: P0 390--405,
P1 347--360, P2 49--52, and P3 8. The retained instrument contains
3,156/1,582/90/8 criteria at these tiers before outcome-specific applicability
checks. The small P2/P3 populations should not be read as broad capability estimates.}
\label{tab:importance-scores}
\begin{center}
\begin{tabularx}{\linewidth}{Yrrrrr}
\toprule
Model & P0 & P1 & P2 & P3 & P0 misses \\
\midrule
GPT-6 Astra & 88.8 & 88.0 & 86.7 & 87.5 & 20.8 \\
GPT-5.6 Terra & 76.5 & 76.9 & 82.2 & 62.5 & 53.3 \\
GPT-5.6 Luna & 76.1 & 74.5 & 72.8 & 50.0 & 58.3 \\
GPT-5.4 & 72.2 & 73.5 & 82.5 & 62.5 & 59.6 \\
GPT-5.2 & 64.9 & 64.7 & 76.0 & 62.5 & 67.6 \\
Claude Opus 5.5 & 70.4 & 70.6 & 62.2 & 50.0 & 41.4 \\
DeepSeek V4-Pro & 65.1 & 63.2 & 65.6 & 75.0 & 72.0 \\
DeepSeek V4-Flash & 62.9 & 62.3 & 70.5 & 75.0 & 78.3 \\
Kimi-K2.5 & 42.2 & 37.8 & 50.0 & 50.0 & 89.0 \\
Grok-4.7 & 80.6 & 79.3 & 79.6 & 75.0 & 33.3 \\
Grok-4.3 & 43.4 & 42.2 & 53.7 & 37.5 & 86.4 \\
Mistral-Large-3 & 21.2 & 18.8 & 17.6 & 0.0 & 96.8 \\
\bottomrule
\end{tabularx}
\end{center}

\end{table}

GPT-6 Astra fulfills 88.8\% of critical criterion weight on average, yet 84 of
404 eligible tasks have a critical miss. The counts are 216/405 for GPT-5.6 Terra,
314/401 for DeepSeek-V4-Flash, 162/391 for Claude Opus 5.5 and 130/390 for Grok-4.7.
These rates depend on the number and content of applicable requirements and
complement, rather than replace, the benchmark scores.

\subsection{Input-File Groups}
\label{app:input-file-analysis}

Tasks are grouped by supplied input types before inspecting model outputs.
Groups overlap because a task may contain several types. Each column reports
task-level Weighted Score, not per-file performance, required output types or
whether the agent inspected every supplied file.

\begin{table}[H]
\caption{Weighted Score by supplied input-file type (\%).
Group sizes are 318/280/250/195/33/55 tasks, respectively. Groups overlap
and retain the main table's declared zeros. Plain-text and structured inputs
occur in every task and would repeat the aggregate; slides have no input
coverage, and heterogeneous residual formats are not a single interpretable group.}
\label{tab:input-file-scores}
\begin{center}
\begin{tabularx}{\linewidth}{Yrrrrrr}
\toprule
Model & Docs & Tables & Code & Images & CAD & Media \\
\midrule
GPT-6 Astra & 85.9 & 87.3 & 85.4 & 83.4 & 90.9 & 78.8 \\
GPT-5.6 Terra & 77.9 & 76.4 & 78.9 & 80.3 & 68.1 & 83.0 \\
GPT-5.6 Luna & 76.7 & 75.7 & 78.4 & 77.2 & 71.4 & 73.7 \\
GPT-5.4 & 74.6 & 74.4 & 74.9 & 76.0 & 66.5 & 68.6 \\
GPT-5.2 & 65.0 & 62.7 & 66.3 & 65.6 & 59.2 & 63.1 \\
Claude Opus 5.5 & 64.9 & 67.3 & 63.0 & 66.4 & 59.2 & 67.1 \\
DeepSeek V4-Pro & 63.3 & 62.2 & 62.0 & 62.5 & 56.9 & 62.8 \\
DeepSeek V4-Flash & 62.7 & 60.5 & 61.9 & 62.5 & 54.8 & 57.5 \\
Kimi-K2.5 & 39.2 & 37.2 & 40.7 & 42.1 & 28.0 & 37.2 \\
Grok-4.7 & 75.9 & 76.3 & 75.5 & 77.3 & 69.9 & 77.6 \\
Grok-4.3 & 42.3 & 41.2 & 43.3 & 45.5 & 30.3 & 41.3 \\
Mistral-Large-3 & 19.7 & 20.1 & 21.4 & 23.7 & 11.8 & 23.7 \\
\bottomrule
\end{tabularx}
\end{center}

\end{table}

GPT-6 Astra and GPT-5.6 Terra reverse order between the CAD and media groups.
Kimi-K2.5~\citep{team2026kimi} scores 40.7 on code-containing tasks and 28.0 on CAD-containing tasks.
These are task-population differences, not isolated file-format effects:
domain, requested work and accompanying inputs vary together.

\subsection{Pass Rates within a Turn Budget}
\label{app:benchmark-budget-results}

\begin{figure}[!htb]
\centering
\includegraphics[width=\linewidth]{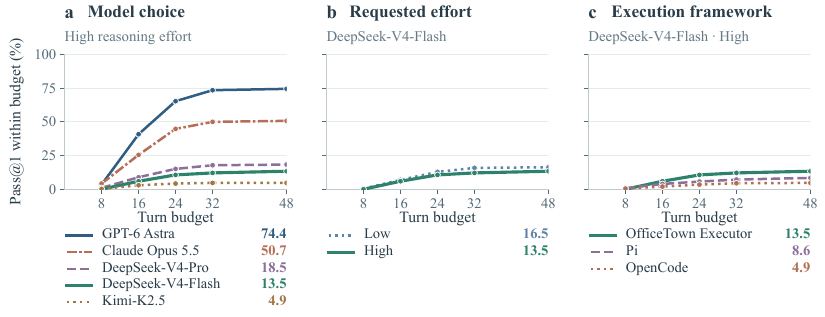}
\caption{Pass@1 at five recorded turn budgets. Populations, selected attempts,
requested settings and turn counts match Figure~\ref{fig:bench-turn-limits}.
All panels use the same \OTBenchmarkTasks{} tasks. A final pass contributes only
after its recorded execution has ended; all other outcomes contribute zero.
Lines guide the eye; no intermediate artifacts are scored.}
\label{fig:bench-turn-limits-pass}
\end{figure}

Figure~\ref{fig:bench-turn-limits-pass} replaces Weighted Score with binary
passes while retaining Figure~\ref{fig:bench-turn-limits}'s records, populations
and ending-turn rule. Model-panel endpoints reproduce
Table~\ref{tab:otbench-fill}; effort and framework panels also cover the complete
benchmark. At 48 turns, Low and High reach 16.5\% and 13.5\% Pass@1; OfficeTown
Executor, Pi and OpenCode reach 13.5\%, 8.6\% and 4.9\%, respectively.
The gap from weighted scores distinguishes partial fulfillment from complete delivery.

\label{app:failure-analysis}

These diagnostics describe associations, not causal failure mechanisms.
Requirement content, artifact usability and failure modes need separate reviewed
annotations; unresolved grading is not evidence of incorrect model work.

\section{Training Data and Model Adaptation}
\label{app:training}
\label{app:training-details}

\subsection{Training Resources}
\label{app:training-resources}

The preparation cohort contains 6,172 cases, including 6,082 with completed
text-consistency editing records. Teacher execution covers 3,718 file-based
tasks; service-dependent tasks lie outside this configuration rather than count
as failed executions. These are preparation counts, not a census of raw Live activity.

The SFT resource contains \OTSFTTrajectories{} complete trajectories distilled from
DeepSeek-V4-Flash~\citep{xu2026deepseek}. From 2,317 trajectories in the original SFT partition, we
exclude 46 tool-serialization cases, 62 exceeding context capacity and seven
parser round-trip cases, leaving 2,202. Trajectories are neither truncated nor
filtered by task-success score. The resource is split into 1,982 training and
220 validation demonstrations. Two training and four validation padding rows
contain no supervised tokens and are not counted as additional demonstrations.

OPD provides \OTOPDTasks{} task--environment pairs with task requests and original
initial environments, without teacher response histories. Its public system
message, chat template and serialized tool definitions
align with the SFT input format. The benchmark and training resources are not
a single random three-way partition of the intermediate construction pool.

\subsection{SFT and On-Policy Distillation}

Qwen3.5-9B is trained first with SFT, then with on-policy distillation~\citep{agarwal2024policy}
using Qwen3.8-Flash-Next as teacher. Table~\ref{tab:training-configuration}
summarizes the resources and teachers; selection and reporting follow
Appendix~\ref{app:data-separation}.

\begin{table}[H]
\caption{Training resources and teachers. Example counts exclude zero-supervision
padding rows; the SFT token total describes the complete demonstration resource.
Optimization settings are given in Tables~\ref{tab:sft-configuration}
and~\ref{tab:opd-configuration}.}
\label{tab:training-configuration}
\begingroup
\begin{center}
\begin{tabularx}{\linewidth}{p{0.37\linewidth}YY}
\toprule
Item & SFT & OPD after SFT \\
\midrule
Initial student & Qwen3.5-9B & SFT-trained Qwen3.5-9B \\
Training inputs & Complete demonstrations & Requests and initial environments \\
Training examples & 1,982 & 828 \\
Validation examples & 220 & None \\
Supervised tokens in resource & 30,704,800 & Not applicable \\
Teacher & DeepSeek-V4-Flash & Qwen3.8-Flash-Next \\
\bottomrule
\end{tabularx}
\end{center}
\endgroup
\end{table}

SFT updates all parameters of Qwen3.5-9B for two epochs using AdamW,
with a 14-step warmup followed by linear learning-rate decay.
Each epoch comprises 248 optimizer steps under the
global batch size of eight, including the zero-supervision padding rows.
The settings in Table~\ref{tab:sft-configuration} preserve complete sequences
without truncation.

\begin{table}[H]
\caption{Full-parameter SFT configuration. The 1,982 training and 220 validation
demonstrations are distinct from the two and four zero-supervision padding rows.
Training uses 248 steps per epoch and 496 steps in total.}
\label{tab:sft-configuration}
\begin{center}
\begin{tabularx}{\linewidth}{p{0.37\linewidth}Y}
\toprule
Parameter & Setting \\
\midrule
Model / training mode & Qwen3.5-9B; full-parameter SFT \\
Hardware / parallelism & One node, $8\times$ NVIDIA A100 80 GB; FSDP2 \\
Schedule / random seed & 2 epochs, 496 steps; seed 42 \\
Global / micro-batch size & 8 / 1 per GPU \\
Optimizer & AdamW; $\beta_1=0.9$, $\beta_2=0.95$ \\
Learning rate & $10^{-6}$; linear decay to $10^{-7}$ (minimum ratio 0.1) \\
Warmup & 3\%; 14 steps \\
Weight decay / gradient clipping & 0.01 / 1.0 \\
Maximum sequence length & 131,072 tokens; no truncation \\
Precision & FP32 master parameters and Adam states; BF16 computation \\
Memory optimizations & Gradient checkpointing, CPU activation offload, LM-head chunk size 512 and length grouping \\
Checkpoint saving & Steps 100, 200, 300, 400 and 496; BF16 Hugging Face model export \\
Validation & At step 496 \\
\bottomrule
\end{tabularx}
\end{center}
\end{table}

OPD samples one student trajectory per task and queries the teacher for the
log probability of each sampled token. For token $a_t$ and context $h_t$,
the distillation signal is
\[
k_{1,t}=\log\pi_{\mathrm{student}}(a_t\mid h_t)
       -\log\pi_{\mathrm{teacher}}(a_t\mid h_t).
\]
The negative log ratio supplies a stop-gradient token-level advantage to a
vanilla policy-gradient update, rather than a directly differentiated supervised
loss. Updates use token-mean aggregation and a distillation coefficient of 1.0.
Task rewards and the separate reference-policy KL term are disabled.

The run uses all 828 training tasks for one epoch: seven batches of 104 and
a final batch of 100. The final batch is retained, with scheduling padding
that does not add task samples. Validation is disabled; no training rows are
relabelled as an independent held-out set. The teacher runs in a separate
vLLM service on the same eight GPUs. Table~\ref{tab:opd-configuration} records
the OPD settings.

\begin{table}[H]
\caption{OPD training configuration. The student is initialized from SFT-trained
Qwen3.5-9B. Length limits are in tokens; the final 100-task batch is retained.}
\label{tab:opd-configuration}
\begin{center}
\begin{tabularx}{\linewidth}{p{0.37\linewidth}Y}
\toprule
Parameter & Setting \\
\midrule
Hardware & One node, $8\times$ NVIDIA B200 \\
Training schedule & One epoch; 8 updates; seed 42 \\
Learning rate & $2.0\times 10^{-6}$ \\
Batch / mini-batch size & 104 / 104; one optimization pass per update \\
Micro-batch size & 1 per GPU \\
Distillation objective & $k_1$; vanilla policy gradient; token-mean aggregation \\
Distillation coefficient / log-probability floor & $1.0$ / $-20.0$ \\
Task rewards / reference-policy KL & Disabled / disabled \\
Rollout correction & Token-level importance sampling; threshold 2.0 \\
Sampling & One trajectory per task; temperature 1.0; top-$p$ 1.0; top-$k$ disabled \\
Prompt / response / context limits & 8,192 / 122,880 / 131,072 \\
Interaction limits & 100 assistant turns; 1,800 s per task; 120 s per tool call \\
Tool-output limit & 65,536; middle truncation \\
Training parallelism & FSDP2 across 8 GPUs; parameter and optimizer offload disabled \\
Rollout execution & Asynchronous vLLM; tensor parallelism 1; 32 sequences; 16 agent workers \\
Rollout GPU-memory fraction & 0.45; prefix caching enabled \\
Memory optimizations & Gradient checkpointing, padding removal and fused Torch kernels \\
Tool-call serialization & \texttt{qwen3\_coder}; released tokenizer, tool schemas and agent loop; strict tokenization checks \\
Checkpointing / validation & Save every 2 updates, retain 2 checkpoints; validation disabled \\
\bottomrule
\end{tabularx}
\end{center}
\end{table}

\subsection{External Evaluation Protocol}
\label{app:training-evaluation}

Base, SFT and SFT + OPD use the OfficeTown Pi/verl execution loop and native tools,
with benchmark-specific scoring and limits.

GDPval excludes tasks requiring live web search. GPT-5.5 scores the original
rubrics as absolute scores, not the official pairwise win rate.

JobBench uses 63 released Easy environments, native Grok-4.3 weighted grades and
a 7,200-second task budget without an additional turn cap. Later recovery phases
exclude verified infrastructure-blocking time from active time.

APEX-Agents v1.1 uses 150 assistant turns and 16,200 seconds per task. Its verifier
uses an Azure DeepSeek-V4-Flash route instead of the published Fireworks snapshot.

Terminal-Bench 2.1 excludes search-dependent tasks and uses native binary rewards.

\clearpage
\subsection{Training Dynamics}
\label{app:training-dynamics}

\begin{figure}[H]
\centering
\includegraphics[width=\linewidth]{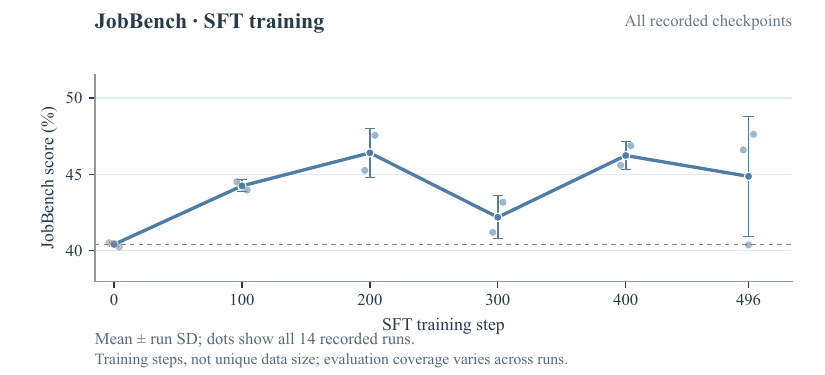}
\caption{JobBench-Easy performance over SFT training. All recorded checkpoints
from the same training run and the Base reference are shown. Points are run
means and error bars are one sample standard deviation; the 14 individual runs
are also displayed. The non-monotonic curve measures training progress, not an
ablation of unique training-data size.}
\label{fig:sft-training-progress}
\end{figure}

Figure~\ref{fig:sft-training-progress} shows non-monotonic SFT progress on
JobBench-Easy, including the decline at step 300 and larger variation at step 496.
Error bars describe recorded run variation, not independent-seed confidence intervals.

\subsection{Budget-Conditioned Outcomes}
\label{app:training-curves}

\begin{figure}[H]
\begin{minipage}[t]{0.49\linewidth}
\centering
\includegraphics[width=\linewidth]{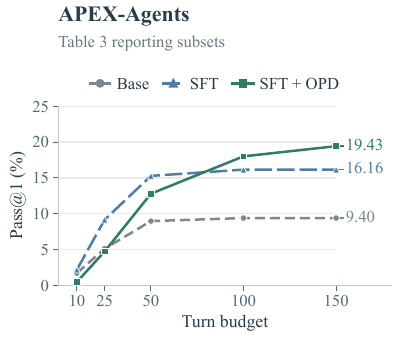}
\caption{Recorded APEX-Agents Pass@1 at five budgets. A pass counts only after
execution ends; endpoints match Table~\ref{tab:training-fill}.}
\label{fig:training-budget-apex}
\end{minipage}\hfill
\begin{minipage}[t]{0.49\linewidth}
\centering
\includegraphics[width=\linewidth]{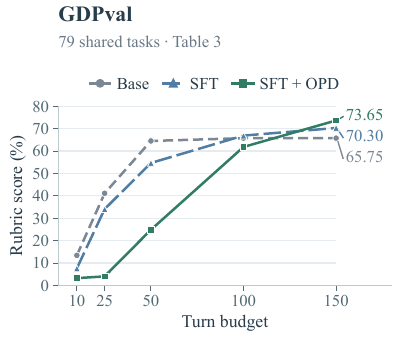}
\caption{Recorded rubric scores on GDPval's shared tasks. SFT + OPD reaches
a higher final score but needs more turns to reach 50 in the full records.}
\label{fig:training-budget-gdpval}
\end{minipage}
\end{figure}

\begin{figure}[H]
\begin{minipage}[t]{0.49\linewidth}
\centering
\includegraphics[width=\linewidth]{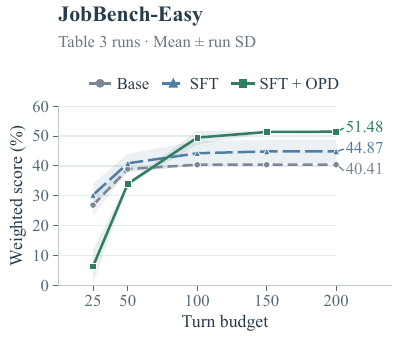}
\caption{Recorded JobBench-Easy scores. Points average original run means
equally; shading is one sample SD across those runs, not an independent-seed
confidence interval.}
\label{fig:training-budget-jobbench}
\end{minipage}\hfill
\begin{minipage}[t]{0.49\linewidth}
\centering
\includegraphics[width=\linewidth]{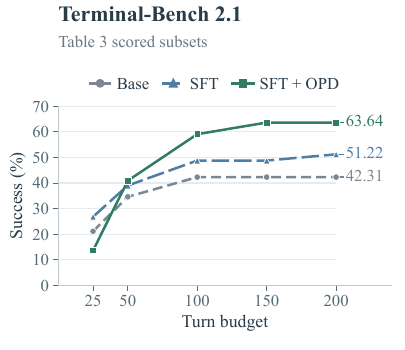}
\caption{Recorded Terminal-Bench 2.1 success rates at five budgets, using
Table~\ref{tab:training-fill}'s selected outcomes and preserving its endpoints.}
\label{fig:training-budget-terminal}
\end{minipage}
\end{figure}

Figures~\ref{fig:training-budget-apex}--\ref{fig:training-budget-terminal} plot
five fixed turn budgets per benchmark; endpoints match Table~\ref{tab:training-fill}.
They summarize existing executions rather than reruns stopped at each budget.
Lines guide the eye between recorded points. Each execution contributes zero
before its recorded ending turn and its final score thereafter, using a fixed
denominator; intermediate artifacts are not rescored. JobBench keeps the original
partial grade when a run stops at the response-token limit and follows the
benchmark's rules for missing grades. Turn counts are not directly comparable as
inference cost.

\end{document}